\documentclass{article}

\usepackage{microtype}
\usepackage{amssymb}
\usepackage{amsmath}
\usepackage{graphicx}
\usepackage{placeins}
\usepackage{multirow}
\usepackage{booktabs} % for professional tables
\usepackage{graphbox}
\usepackage{tcolorbox}
\usepackage{paralist}
\usepackage{soul}
\usepackage{pifont}
\usepackage{tabularx}
\usepackage{subfig}
\usepackage{verbatim}
\usepackage{float}
\usepackage{wrapfig}
\usepackage{array}

\DeclareMathOperator*{\argmin}{\arg\!\min}
\newcommand\norm[1]{\left\lVert#1\right\rVert}
\usepackage[utf8]{inputenc} % allow utf-8 input
\usepackage[T1]{fontenc}    % use 8-bit T1 fonts
\usepackage{hyperref}       % hyperlinks
\usepackage{url}            % simple URL typesetting
\usepackage{amsfonts}       % blackboard math symbols
\usepackage{nicefrac}       % compact symbols for 1/2, etc.

\usepackage[accepted]{arxiv}
\usepackage{xspace}
\usepackage{booktabs}
\newcommand{\ra}[1]{\renewcommand{\arraystretch}{#1}}

\icmltitlerunning{Calibration and Uncertainty Quantification of Bayesian Neural Networks for Geophysical Applications}

\begin{document}

\twocolumn[
\icmltitle{Calibration and Uncertainty Quantification of \\ Bayesian Convolutional Neural Networks for Geophysical Applications}

\begin{center}
    \begin{tabular}{cccc}
         \textbf{Lukas Mosser} \& \textbf{Ehsan Zabihi Naeini} \\
         Earth Science Analytics
    \end{tabular}
\end{center}
\vskip 0.3in

\begin{abstract}
\vspace*{10px}
Deep neural networks offer numerous potential applications across geoscience, for example, one could argue that they are the state-of-the-art method for predicting faults in seismic datasets. In quantitative reservoir characterization workflows, it is common to incorporate the uncertainty of predictions thus such subsurface models should provide calibrated probabilities and the associated uncertainties in their predictions. It has been shown that popular Deep Learning-based models are often miscalibrated, and due to their deterministic nature, provide no means to interpret the uncertainty of their predictions. We compare three different approaches to obtaining probabilistic models based on convolutional neural networks in a Bayesian formalism, namely Deep Ensembles, Concrete Dropout, and Stochastic Weight Averaging-Gaussian (SWAG). These methods are consistently applied to fault detection case studies where Deep Ensembles use independently trained models to provide fault probabilities, Concrete Dropout represents an extension to the popular Dropout technique to approximate Bayesian neural networks, and finally, we apply SWAG, a recent method that is based on the Bayesian inference equivalence of mini-batch Stochastic Gradient Descent. We provide quantitative results in terms of model calibration and uncertainty representation, as well as qualitative results on synthetic and real seismic datasets. Our results show that the approximate Bayesian methods, Concrete Dropout and SWAG, both provide well-calibrated predictions and uncertainty attributes at a lower computational cost when compared to the baseline Deep Ensemble approach. The resulting uncertainties also offer a possibility to further improve the model performance as well as enhancing the interpretability of the models.
\end{abstract}]

\section{Introduction}\label{sec:introduction}
Interpretation of geophysical data, and more specifically seismic data, is often time-consuming, subjective, and labor-intensive. The advent of Deep Learning \citep{lecun2015deep}, and its successful applications to automate many computer vision tasks \citep{krizhevsky2012imagenet,russakovsky2015imagenet}, have allowed for significant advances in the ability to efficiently detect geological structures from 3D seismic volumes \citep{pham2018automatic,wu2019faultseg3d}. The interpretation of faults within a geological basin or reservoir from seismic data is one of such tasks that is also associated with high uncertainties \citep{lecour2001modelling,thore2002structural}. Recently, numerous approaches using machine learning have been presented to automate the process of identifying fault planes within seismic images \citep{Pochet2019SeismicFD}, which appear to outperform traditional fault detection techniques such as attribute-based methods \citep{chopra2007volumetric, hale2013methods,qi2019image}. Due to the lack of labeled data, it has been recently proposed to use synthetic generated training data to simulate seismic images with artificial representations of the geological features of interest \citep{wu2019building,gao2020channel}. 

For many practical applications of these methods, it is not sufficient to predict the presence of a geological structure, but more importantly one would like to also know the associated uncertainties of a specific prediction. This is not unique to subsurface studies using geophysical data but is generally a critical step in many high-risk settings. For example in medical diagnostic applications, the model predictions may determine patients’ health with the right course of treatment. Self-driving cars are another example where the model's uncertainty of detecting a pedestrian could be used to slow down and eventually switch to manual driving \citep{kendall2017uncertainties}. In hydrocarbon exploration, uncertainty quantification could mean the difference between a dry hole or a discovery. As an example, there could be an overlap between the elastic properties of soft shale rock and those of a gas sand reservoir. Assume we have a model that is trained to predict different rock and fluid types and has a-priori knowledge of both soft shale and gas sand classes. The uncertainty of the final predictions can be key in de-risking the target reservoir as one scenario might be that the model predicts a gas sand body but with high uncertainty. Similarly, in the case of incomplete a-priori knowledge e.g. where the model is only aware of a gas sand type, it is as important to measure and capture the uncertainty. 

However, in the case of Deep Learning applications for seismic data, there are challenges in measuring the probabilities and uncertainties of the outcomes. \citet{guo2017calibration} have shown that the probabilities of deep neural networks (DNNs) are poorly calibrated, and may significantly over- or under-estimate the associated probabilities of the detected geological features. While the proposed methods can have a good performance in detecting the target features, many of these approaches do not allow investigation of the associated uncertainties \citep{pham2018automatic,mosser19uncertainty} that arise in the object-detection process. As mentioned before, such uncertainties may have a critical impact on the decision-making process that one might have to deal with in exploration and field development settings \citep{thore2002structural, mosser19uncertainty}. Tackling the problem of probability calibration and uncertainty estimation for deep neural networks allows incorporating these methods in structural interpretation and reservoir characterization workflows and reduces the risk in the decision-making process.

There have been some attempts, however, to address these shortcomings and our contribution can be seen as a further development of the previous work by \citet{pham2018automatic} and \citet{mosser19uncertainty}
and a direct extension to the work by \citet{mosser2020deep} and \citep{feng2021uncertainty}. In contrast to the latter two most recent papers, we extend the evaluation of Deep Learning-based probabilistic methods to the problem of model calibration. Furthermore, we conduct a thorough comparison of different approaches to obtain probabilistic methods which can be formulated in a Bayesian framework \citep{wilson2020bayesian}. 

We first introduce a framework for obtaining probabilistic models based on deep neural networks, and in our specific use case deep convolutional neural networks (CNNs), in a Bayesian framework. Subsequently, we show the different types of associated uncertainties involved in evaluating probabilistic models. We then introduce three approaches to obtain well-calibrated probabilistic models based on deep neural networks; Deep Ensembles \citep{lakshminarayanan2017simple} which combine the predictions from separately trained deep neural networks, Concrete Dropout \citep{gal2017concrete} - an extension to the popular Dropout approach \citep{gal2016dropout} to obtain uncertainty representations which requires no hyper-parameter tuning of the associated Dropout probabilities, and finally Stochastic Weight Averaging - Gaussian (SWAG) \citep{maddox2019simple}, which builds on the concept of Stochastic Gradient Descent (SGD) as an approximate Bayesian inference method \citep{mandt2017stochastic}. 

We compare these three probabilistic approaches to deterministic counterparts commonly used in seismic object detection and segmentation applications (here we choose seismic fault detection). We provide quantitative estimates of model calibration and uncertainty estimation on synthetic datasets, and their robustness to out-of-distribution artifacts is evaluated. The probabilistic methods are then applied to real seismic datasets with varying seismic imaging qualities and signal-to-noise ratio. The probability attributes and their associated uncertainty estimates are presented, and the impact of different stochastic realizations on geo-modeling workflows are highlighted based on an example of fault-stick extraction.

\section{Bayesian Learning and Neural Networks}
\label{sec:bayesian_learning}
Deep Neural Networks can be represented as layers of consecutive linear operators i.e. neurons applied on input data $\mathbf{x}$ combined with a non-linear transformation i.e. activation function. They can be considered as general models or function approximators \citep{cybenko1989approximation} which allow one to approximate any given function to arbitrary precision by increasing the number of intermediate representations; that is by increasing the number of layers and the number of neurons in each layer. As such, DNNs can arguably capture complex relationships given a large number of model parameters i.e. the weights $\mathbf{W}$ and biases $\mathbf{b}$ of each neuron in the network. Learning the optimal setting of these model parameters can be achieved in a supervised setting, where for each input $\mathbf{x}$ an associated $\mathbf{y}$ value is provided; $\mathbf{y}$ can be a continuous property, often referred to as \textit{target set} and hence a regression problem, or a discrete property often referred to as \textit{label set} and therefore a classification problem.

While neural networks have found applications in other learning schemes, such as unsupervised or reinforcement learning, we will focus on supervised classification problems only using CNNs which are a sub-category of DNNs in which the network itself generates the required attributes via convolutional layers acting as filters on the input image \citep{fukushima1982neocognitron,lecun1989backpropagation}.  We will refer to the data used to learn the parameters of the models as the training dataset $\mathcal{D}={(\mathbf{x}, \mathbf{y})}$. Once the neural network has been optimized on the training dataset, we use a portion of the data called the test dataset $\mathcal{D}^*={(\mathbf{x}^*, \mathbf{y}^*)}$ unseen by the network to assess the performance.
\newpage
We begin with constructing the objective function which forms the foundation of the learning process. In a probabilistic sense, the likelihood of a label $\mathbf{y}_i$ for a given input sample $\mathbf{x}_i$ is represented via the conditional probability
\begin{equation}
\begin{gathered}
    p(\mathcal{D}|\mathbf{\omega})=p(\mathbf{y}|\mathbf{x}, \mathbf{\omega}) = \prod_{i=1}^N p(y_i|x_i, \mathbf{\omega})\\ 
    =\prod_{i=1}^N \text{Pr}(Y=y_i|X=x_i)\\
    =\prod_{i=1}^N p(x_i; \mathbf{\omega})^{y_i}(1-p(x_i; \mathbf{\omega}))^{(1-y_i)}
\end{gathered}\label{eq:likelihood}
\end{equation}
where $N$ is the total number of samples. We assume that we represent the probability of the positive class $\text{Pr}(Y=1|X=x_i)$ for an input sample $x_i$ in a binary classification problem by a model $p(x_i; \mathbf{\omega})$ that depends on the input and a set of weights $\mathbf{\omega}$. We assume a Bernoulli-likelihood function and for numerical reasons take the logarithm of the likelihood transforming the product in Equation~\ref{eq:likelihood} into a sum
\begin{equation}
\begin{gathered}
    \log p(\mathcal{D}|\mathbf{\omega})=\log p(\mathbf{y}|\mathbf{x}, \mathbf{\omega})\\ =\sum_{i=1}^N y_i\log p(x_i; \mathbf{\omega})+(1-y_i)\log \left(1-p(x_i; \mathbf{\omega})\right)
    \end{gathered}\label{eq:loglikelihood}
\end{equation}
which can be maximized to obtain the so-called maximum-likelihood estimate of the model parameters given the training data. The log-likelihood expressed in Equation~\ref{eq:loglikelihood} can be considered as an objective function as it solely depends on the parameters $\mathbf{\omega}$. The learning process is then optimized by minimizing this objective function commonly referred to as the loss by summing the \textit{negative} log-likelihood computed over the entire training dataset
\begin{subequations}
\begin{equation}
    \mathcal{L}(\mathbf{\omega}) = - \sum_{i=1}^N y_i\log p(x_i; \mathbf{\omega})+(1-y_i)\log \left(1-p(x_i; \mathbf{\omega})\right)\label{eq:loss}
\end{equation}
\begin{equation}
    \mathbf{\omega}_{MLE}=\argmin_{\mathbf{\omega}} \mathcal{L}(\mathbf{\omega})\label{eq:minimize_loss}
\end{equation}
\end{subequations}
where $\mathbf{\omega}_{MLE}$ is the maximum-likelihood estimate of the model parameters. Depending on the parameterization of the model, simply minimizing Equation~\ref{eq:loss} on the training dataset can lead to overfitting and therefore poor generalization of the model $p(x_i; \mathbf{\omega})$ to new unseen data $\mathbf{x}^*$. There are different strategies to mitigate this problem. In a probabilistic framework, one can estimate the distribution of the weights instead of the most likely solution and furthermore, instead of being agnostic about the choice of weights, which can in theory take on any real value, incorporate some well-defined distribution referred to as a prior. Taking into account these prior assumptions and the likelihood (Equation ~\ref{eq:likelihood}) allows us to define a probabilistic model 
\begin{equation}
    \begin{gathered}
    \mathbf{\omega}\sim p(\mathbf{\omega}) = \mathcal{N}(0, \sigma\mathbf{I}) \\
    \hat{p}=NN(\mathbf{x}, \mathbf{\omega})
\end{gathered}
\label{eq:probabilistic_model}
\end{equation}
where we have assumed a neural network model ($NN$) to represent the predicted probability of the positive i.e. fault class, parameterized by weights $\mathbf{\omega}$ following a Gaussian prior distribution with variance $\sigma^2$.

Bayes' law provides a mathematical machinery to combine the observed training data with our prior modeling assumptions i.e. the choice of model being a neural network, and the prior distribution over its weights
\begin{subequations}
\begin{equation}
    p(\mathbf{\omega}|\mathcal{D}) = \frac{p(\mathcal{D}|\mathbf{\omega})p(\mathbf{\omega})}{p(\mathcal{D})}\label{eq:bayes}
\end{equation}
\begin{equation}
    p(\mathcal{D}) = \int p(\mathcal{D}|\mathbf{\omega})p(\mathbf{\omega})d\mathbf{\omega} \label{eq:evidence}
\end{equation}
\end{subequations}
(Equation~\ref{eq:bayes}) allows us to update our assumptions of what model parameters best represent the data which is expressed as the so-called posterior distribution $p(\mathbf{\omega}|\mathcal{D})$. In practice, to obtain the posterior probability, given the likelihood and prior probabilities, the so-called evidence $p(\mathcal{D})$ (Equation~\ref{eq:evidence}) is treated as a constant factor in Equation~\ref{eq:bayes} (it is intractable to compute the evidence for models with a large number of parameters). Thus maximizing the posterior probability leads to
\begin{equation}
\begin{gathered}
 \mathbf{\omega}_{MAP} = \underset{\mathbf{\omega}}{\mathrm{argmax}} \left\{ \log p(\mathcal{D}|\mathbf{\omega}) + \log p(\mathbf{\omega})\right\}\\=\underset{\mathbf{\omega}}{\mathrm{argmin}}  \left\{ \mathcal{L}(\mathbf{\omega}) - \log p(\mathbf{\omega})\right\}
\end{gathered}\label{eq:map} 
\end{equation}
which is commonly referred to as Maximum A-Posteriori (MAP) estimation of the model parameters. It can be observed that here the prior $p(\mathbf{\omega})$  acts as a regularisation. When the model is a deep neural network, we can solve Equation~\ref{eq:map} iteratively using Backpropagation \citep{linnainmaa1976taylor,rumelhart1986learning} and gradient-descent-based minimization of the objective function. This allows finding one set of weights that correspond to a single mode of the posterior distribution (Equation~\ref{eq:bayes}). 

When considering the Bayesian formulation of the learning problem, the prior, the likelihood, and the posterior are represented by a probability distribution. Rather than estimating one set of weights (Equation~\ref{eq:map}), the posterior distribution of the model weights is estimated. This then allows one to capture the uncertainty of predictions by the model on new data due to uncertainty in the model weights. To achieve this, instead of optimizing the network directly for one set of weights, an average over all possible weights is computed via a process known as marginalization
\begin{equation}
    p(\mathbf{y}^*|\mathbf{x}^*, \mathcal{D})=\int_{\Omega}p(\mathbf{y}^*|\mathbf{x}^*, \mathbf{\omega})p(\mathbf{\omega}|\mathcal{D})d\mathbf{\omega}
    \label{eq:posteriorpredictive}
\end{equation}
where $\Omega$ is the neural network weight space represented by $p(\mathbf{\omega})$ and $(\mathbf{x}^*, \mathbf{y}^*)$ correspond to new unseen data for which we want to infer $p(\mathbf{y}^*|\mathbf{x}^*)$. One can see that instead of taking a single set of weights that maximizes the probability of the posterior distribution, all possible weights are being considered and subsequently weighted by their probability. We can think of this as an ensemble of models weighted by the probability of each model. Indeed this is equivalent to an ensemble of an infinite number of neural networks, with the same architecture but with different weights. 

This integral appears in many cases of Bayesian learning, however, the solution to it remains intractable. This is because the posterior probability $p(\mathbf{\omega}|\mathcal{D})$ cannot be computed analytically.  There are various approaches that have been developed over the years to overcome this. Sampling methods approximate it with the average of samples drawn from the posterior distribution. One way to do that is the Markov-Chain Monte-Carlo-based methods \citep{mackay1992practical,neal2012bayesian}. Another solution is to approximate the true intractable distribution with a different distribution from a tractable family where the similarity of the two distributions can be measured via Kulback-Leibler (KL) divergence \citep{lee2020variational}.

Once we have obtained an approximation to the true posterior, we characterize the distribution (Equation~\ref{eq:posteriorpredictive}) by its moments. First, we estimate the mean by drawing samples from the posterior weight distribution $\mathbf{\omega}\sim p(\mathbf{\omega}|\mathcal{D})$ which is practically often feasible using a Monte-Carlo estimate
\begin{equation}
\begin{gathered}
\mathbb{E}[p(\mathbf{y}^*|\mathbf{x}^*, \mathcal{D})]=\frac{1}{T}\sum_{t}^T p(\mathbf{y}^*|\mathbf{x}^*, \mathbf{\omega_t}), \\ \text{where} \ \mathbf{\omega_t}\sim p(\mathbf{\omega}|\mathcal{D})
\end{gathered}\label{eq:montecarlomean}
\end{equation}
where $T$ is the number of Monte-Carlo samples obtained to estimate the first-moment (mean) of the posterior predictive distribution (Equation~\ref{eq:posteriorpredictive}).

\begin{figure*}[!htb]
   \centering
\includegraphics[width=\textwidth]{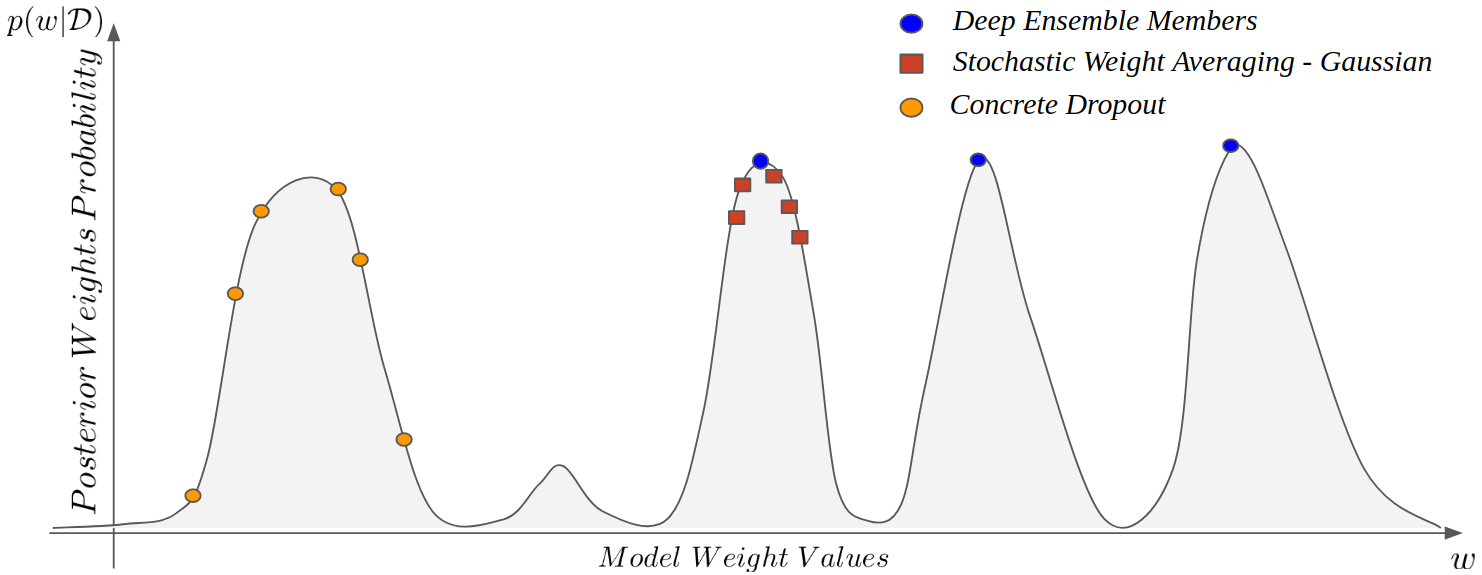}
   \caption{The posterior of the CNN weights given the training data is multi-modal. When marginalizing to obtain the posterior-predictive distribution, 
   different methods represent different regions of the posterior. Deep Ensembles consider only the MAP solutions close to individual modes, while techniques such as Stochastic Weight Averaging-Gaussian or Concrete Dropout explore the vicinity of a single mode. Figure after \citet{wilson2020bayesian}}\label{fig:posteriorpredictive}
\end{figure*}

Due to the multi-modal nature of the posterior distribution over the model weights (Figure~\ref{fig:posteriorpredictive}), sampling from this distribution needs to be computationally fast, while allowing one to explore a wide range of the posterior distribution to obtain a good characterization of the posterior predictive distribution (Equation~\ref{eq:posteriorpredictive}). To further characterize the posterior, we of course consider the second moment (variance) of this predictive distribution and hence the associated uncertainty of the model predictions. There are different types of uncertainty, and each is useful for different purposes \citep{zeldes2017deep}. In general, we can decompose the total uncertainty into two parts: the model uncertainty, and the data uncertainty.

Model uncertainty, also known as epistemic uncertainty, accounts for uncertainty in the ability of our model to explain the provided data. Given more data the epistemic uncertainty decreases. This type of uncertainty is important in high-risk applications and when the training dataset is small. As an example, when considering improvements to an existing model, characterization of the epistemic uncertainty could help to identify the regions where additional data should be collected.

Data uncertainty, or aleatoric uncertainty, captures the noise inherent in the observation. Obtaining more data will not help in that case, because the noise is inherent in the data. For example, the noise and resolution limits imposed on seismic data measurements due to acquisition setup are inherent in the data and there is no physically meaningful remedy for it. 

\citet{kwon2018uncertainty} and \citet{depeweg2019modeling} provide different representations of how the total, aleatoric, and epistemic uncertainties can be evaluated. Here, we follow the derivation by \citet{depeweg2019modeling} where the entropy $\mathcal{H}(\mathbf{y}^*|\mathbf{x}^*)$ is used to characterize the total uncertainty, and $\mathbb{E}_{q(\mathbf{\omega}|\mathcal{D})}\left[\mathcal{H}(\mathbf{y}^*|\mathbf{x}^*, \mathbf{\omega})\right]$ describes the aleatoric uncertainty. The total uncertainty additively decomposes into the aleatoric and epistemic uncertainties, subtracting the aleatoric from the total uncertainty gives the epistemic uncertainty, which is equivalent to the so-called mutual information
\begin{equation}
 \mathcal{I}(\mathbf{y}^*, \mathbf{\omega}) = \mathcal{H}(\mathbf{y}^*|\mathbf{x}^*)-\mathbb{E}_{p(\mathbf{\omega}|\mathcal{D})}\left[\mathcal{H}(\mathbf{y}^*|\mathbf{x}^*, \mathbf{\omega})\right]\label{eq:mutual_information}
\end{equation}
In practice, this means that we first compute the entropy of the Monte-Carlo estimate of the mean of the marginal distribution (first term in Equation~\ref{eq:mutual_information}) and compute the mean of the entropy for individual predictions (second term in Equation~\ref{eq:mutual_information}). Since we are dealing with a Bernoulli-likelihood (recall Equation~\ref{eq:likelihood}), the entropy of a Bernoulli-variable used to represent pixel-wise predictions of a binary classification (fault vs non-fault in our example) is given by 
\begin{equation}
    \mathcal{H}(p) = -p\log(p) - (1-p)\log(1-p)
\end{equation}
where $\log$ is the natural logarithm, and $p$ in our case are the pixel-wise probabilities for a new data point $\mathbf{x}^*$. However, to incorporate probabilistic models such as probabilistic neural networks and their associated uncertainties in a decision-making process, such models ought to be calibrated, which means that the predicted probability $\hat{p}$ for an event should correspond to its true probability of occurrence $p$.

\citet{guo2017calibration} have shown that deep neural networks commonly used in practical applications are often not calibrated and hence may significantly over- or under-estimate the probability of an event. Furthermore, in many practical geoscience applications, such as fault detection in seismic images, where only a few percent of the samples in a seismic volume can be associated with the presence of a fault, there is often a significant imbalance in the occurrence of class labels. This imbalance can lead to difficulties in training a neural network to detect the minority class, which in our case are the fault labels \citep{BUDA2018249}. To overcome this, \citet{wu2019faultnet} propose to use a balanced binary cross-entropy loss (BCE) which re-balances the contribution of the fault and background labels.

\label{sec:methodology}

\begin{table*}
\centering
\ra{1.3}
\begin{tabularx}{\textwidth}{@{}c>{\raggedright\arraybackslash}X@{}}
\toprule
			$\bold{\large{Method}}$ & \centering$\bold{\large{Description}}$\tabularnewline

			\midrule
			$\bold{Unbalanced-CE}^*$ & A single CNN is trained using SGD with a BCE-loss and $L_2$-weight regularization corresponding to a single MAP estimate of the model weights.\\
			$\bold{Balanced-CE}^*$ & A single CNN is trained using SGD with a balanced BCE-loss \citep{wu2019faultnet}, and $L_2$-weight regularization.\\
			$\bold{Deep \, Ensemble}$ & Three CNNs are trained from random initialization with a BCE-loss, $L_2$-weight regularization, each corresponding to a mode of the weight posterior. Predictions from each model are computed separately and averaged to obtain a prediction of the deep ensemble.\\
			$\bold{Concrete \, Dropout}$ & A single CNN is trained with Concrete Dropout applied to the filter kernels of the convolutional layers. Dropout probabilities are optimized as part of the training process. Predictions are made by applying Dropout with optimized Dropout probabilities at test time and subsequently averaged. \\
			$\bold{SWAG}$ & A single CNN is trained with SGD, a BCE-loss, and $L_2$ regularization, where training the model is considered a stochastic process. The stationary distribution is Gaussian over the model weights $\mathbf{\omega}$. A low-rank approximation of the Gaussian over consecutive SGD-iterates is formed. After training, weights are sampled from this Gaussian, and predictions averaged over a number of realizations of the weights. \\
\bottomrule
\end{tabularx}
\caption{Description of the different considered methods used to create deterministic (*) and probabilistic deep neural networks}\label{tab:methods}
\end{table*}

\section{Methodology}
As previously highlighted, solving the Bayesian inference problem when considering models based on deep neural networks is intractable (Equation~\ref{eq:bayes}). Nevertheless, there are numerous methods to obtain approximate solutions and to subsequently characterize the posterior predictive distribution (Equation~\ref{eq:posteriorpredictive}). We consider three recently developed approaches as summarized in Table~\ref{tab:methods} to find approximate solutions to this challenging inference problem, which will be subsequently applied to the task of fault detection in seismic images.

One approach to obtaining well-calibrated models is to use a proper scoring rule \citep{gneiting2007strictly} as an objective function. Proper scoring rules provide a measure for the quality of the predictive uncertainty, and models that minimize the error with respect to a proper scoring rule in turn also provide well-calibrated outputs. In the following, we outline a number of methods to obtain approximate representations of the posterior distribution of the probabilistic model outlined in Equation~\ref{eq:probabilistic_model}. One common scoring rule is the negative log-likelihood which for the case of a binary classification problem is equivalent to the BCE. The task of detecting faults in seismic images is framed here as a binary classification problem with a Bernoulli-likelihood. Therefore, all the approximate posterior inference methods directly optimize the log-likelihood (~\ref{eq:loglikelihood}), and in some cases have additional regularization terms. These approaches are therefore incentivized to produce well-calibrated probability distributions and we evaluate their calibration empirically using a number of synthetic examples. 

\subsection{Deep Ensembles}\label{sec:deep_ensembles}
Mixture models of deep neural networks are a simple but computationally expensive baseline to create probabilistic calibrated predictive models. Here we consider a mixture model based on a single neural network architecture where each model provides probabilities $p(\mathbf{y}^*|\mathbf{x}^*, \mathbf{\omega}_i)$
where $\mathbf{\omega}_i$ are the weights and biases of the i-th neural network of the mixture.
We obtain an estimate of the expectation of the posterior predictive distribution analogous to Equation~\ref{eq:montecarlomean} by an unweighted average of the mixture member's predictions
\begin{equation}
    \mathbb{E}[p(\mathbf{y}^*|\mathbf{x}^*, \mathcal{D})]\approx\frac{1}{M}\sum_{i}^M p(\mathbf{y}^*|\mathbf{x}^*, \mathbf{\omega_i})\label{eq:ensemble}
\end{equation}
where each configuration of the model weights are obtained by complete training on the dataset $\mathcal{D}$ using a MAP objective to find a weight configuration close to a mode of the posterior $p(\mathbf{\omega}|\mathcal{D})$ (Figure~\ref{fig:posteriorpredictive}).

In practice, we train $M$ deep convolutional neural networks with different initializations of the weights $\mathbf{\omega}$ using a BCE-loss function and $L_2$ regularization. This corresponds to obtaining point-based MAP estimates from the multi-modal posterior distribution of the weights given the training data $p(\mathbf{\omega}|\mathcal{D})$. We marginalize over the posterior weight distribution by the Monte-Carlo estimate in Equation~\ref{eq:montecarlomean} using a pixel-wise average of the predictions from each trained neural network and obtain the associated uncertainties (Equation~\ref{eq:mutual_information}) from the individual ensemble member predictions.

\subsection{Concrete Dropout}\label{sec:concrete_dropout}
As outlined in the previous section, solving Equation~\ref{eq:posteriorpredictive} analytically is intractable in high-dimensional model space and therefore approximate Bayesian inference techniques must be used to obtain a representation of the posterior $p(\mathbf{\omega}|\mathcal{D})$. 

\citet{gal2016dropout} have shown that Dropout \citep{srivastava2014dropout} can be interpreted as a Variational Inference (VI) technique to approximate the posterior for Bayesian neural networks. We perform VI using Dropout sampling by first defining the distribution $\mathbf{\omega}\sim q(\mathbf{\omega})$ over the weights. This allows approximation of the true posterior distribution $p(\mathbf{\omega}|\mathcal{D})$ by minimizing the KL-divergence between the variational distribution $q(\mathbf{\omega})$ and the true posterior
\begin{equation}
    \argmin_{\mathbf{\omega}}\textrm{KL}\{q(\mathbf{\omega})||p(\mathbf{\omega}|\mathcal{D})\} 
\end{equation}
\citet{gal2016dropout} have shown that minimizing the KL-divergence is equivalent to maximizing the so-called Evidence Lower Bound (ELBO). In the case of the Dropout variational distribution, this is equivalent to minimizing a BCE-loss with $L_2$ weight regularization which can be implemented conveniently using well-known mini-batch stochastic gradient descent and neural network Backpropagation methods. The approach allows us to solve the intractable Bayesian inference problem presented in Equation~\ref{eq:posteriorpredictive} such that it is applicable to large datasets and deep neural networks.
\begin{comment}
%By using a variational approach we try to find an approximate representation of the posterior distribution that is close to the true posterior. By using a parametric family of distributions $q_{\mathbf{\omega}}(\theta)$ we can reformulate the Bayesian inference problem for the posterior weight distribution of a deep neural network as an optimization task.

After the parameters of the Bayesian neural network have been obtained, we characterize the
predictive distribution analogous to Equation~\ref{eq:montecarlomean} with a Monte-Carlo estimate using samples $\mathbf{\omega}_t\sim q(\mathbf{\omega})$ 
\begin{equation}
\begin{gathered}
    p(\mathbf{y}^*|\mathbf{x}^*) = \int_{\Omega}{p(\mathbf{y}^*|\mathbf{x}^*,\mathbf{\omega})q(\mathbf{\omega})}d\mathbf{\omega}\\
    \approx\frac{1}{T}\sum_{t=1}^T p(\mathbf{y}^*|\mathbf{x}^*,\mathbf{\omega}_t)
\end{gathered}
\end{equation}
where $q(\mathbf{\omega})$ is the so-called variational distribution.

\end{comment}

In their seminal work \citet{gal2016dropout} use a Bernoulli distribution to randomly cancel out weight contributions and weight matrices sampled from a Gaussian prior distribution and to represent the variational Dropout distribution $\mathbf{q}(\mathbf{\omega})$
\begin{subequations}
\begin{equation}
    q(\mathbf{\omega}) \sim \mathbf{W}_i \cdot \text{diag}([z_{i, j}]_{j=1}^{K_i})
    \label{eq:var_q}
\end{equation}
\begin{equation}
\begin{gathered}
    z_{i, j} \sim \text{Bernoulli}(p_i), \\ 
    \text{for} \ i=1,...,L \ \text{and} \ j=1,...,K_i
\end{gathered}\label{eq:var_bernoulli}
\end{equation}\label{eq:variational}
\end{subequations}
where $L$ is the total number of layers in the network, $\mathbf{W}_i$ are the convolutional filter weights, and $K_i$ are the number of convolutional filters in the i-th layer. Multiplying the weight matrices with the Bernoulli-distributed random variable induces a prior over the weight matrices where at random for each realization of $q(\mathbf{\omega})$, on average only a fraction $p_i$ of the i-th layer's convolutional filters contribute to the prediction of the CNN. From a probabilistic point of view, the variational distribution over the weights induced by the Dropout distribution is highly multi-modal (Figure~\ref{fig:posteriorpredictive}). Stochastic VI can be performed by stochastic gradient descent where the objective function to be minimized is defined as:
\begin{subequations}
\begin{equation}
\begin{gathered}
    \mathcal{L}_{MC}(\mathbf{\omega}) = -\frac{1}{B}\sum_{i\in\mathcal{D}}\log p(\mathbf{y}_i|\mathbf{x}_i, \mathbf{\omega})\\
    + \frac{1}{N} \text{KL}\left\{q_{\mathbf{\theta}}(\mathbf{\omega})||p(\mathbf{\omega})\right\}
    \end{gathered}\label{eq:mcdropout_loss}
\end{equation}
\begin{equation}
    \text{KL}\left\{q_{\mathbf{\theta}}(\mathbf{\omega})||p(\mathbf{\omega})\right\} = \sum_{i=1}^{L} \text{KL}\left\{q_{\mathbf{\theta}_i}(\mathbf{\omega}_i)||p(\mathbf{\omega}_i)\right\}
\end{equation}
\begin{equation}
   \text{KL}\left\{q_{\mathbf{\theta}}(\mathbf{\omega})||p(\mathbf{\omega})\right\} \propto \frac{l^2(1-p_i)}{2}\norm{\mathbf{M}}^2-K_i\mathcal{H}(p_i) \label{eq:mcdropout_kl}
\end{equation}
\begin{equation}
   \mathcal{H}(p_{i}) = -p_i\log p_i - (1-p_i)\log(1-p_i) \label{eq:mcdropout_entropy}
\end{equation}
\end{subequations}
where $B$ is the batch-size, $N$ being the total number of elements in the dataset $\mathcal{D}$, $K_i$ are the number of convolutional filters in the i-th layer, and $l$ is the prior length-scale. 

The objective function for Concrete Dropout represented by Equation~\ref{eq:mcdropout_loss} consists of two terms; the first corresponds to the likelihood, which here is represented by a Bernoulli-likelihood parameterized by a CNN with parameters $\mathbf{\omega}$. The second part of Equation~\ref{eq:mcdropout_loss} corresponds to the KL-divergence between the approximate posterior distribution of the parameters of the neural network  $q_{\theta}(\mathbf{\omega})$ and the prior of the network parameters $p(\mathbf{\omega})$. The KL-divergence balances the likelihood so that the approximate posterior stays close to the prior.

For the Monte-Carlo dropout strategy, the additional entropy term in Equation~\ref{eq:mcdropout_kl} remains constant since the objective function (Equation~\ref{eq:mcdropout_loss}) is not optimized with respect to the dropout probabilities. In the case of Concrete Dropout, the KL-divergence (Equation~\ref{eq:mcdropout_kl}) and the entropy regularise the magnitude of the dropout probabilities. Minimizing the KL-divergence for a Bernoulli random variable maximizes the entropy and hence in the limiting case increases the dropout probability to 0.5 \citep{gal2017concrete}. When the number of data N is large, the importance of the KL-divergence in the overall objective function becomes small, and the dropout probabilities decay to zero. The additional scaling with respect to the number of convolutional filters in each layer $K_i$ ensures that there is a trade-off between the number of parameters of the network and the size of the datasets they are trained on. The prior length-scale $l$ is typically combined into an empirically chosen hyper-parameter for weighting the KL-divergence (Equation~\ref{eq:mcdropout_kl}).

The posterior predictive distribution is characterized by obtaining a Monte-Carlo estimate as given in Equation~\ref{eq:montecarlomean}. In practice, this means that we make $T$ predictions at test time while applying Dropout and average the output of the deep neural network. The number of Dropout predictions used to make the Monte-Carlo approximation to the posterior predictive distribution can be evaluated by observing the variance of the mean for increasing numbers of $T$ as is also shown by \citep{feng2021uncertainty}. Once a certain number of weight realizations have been averaged, the mean and variance of the predictions stabilize and the number of predictions $T$ after which stabilization occurs can be used to make predictions and evaluate the associated uncertainties (Equation~\ref{eq:mutual_information}) on new unseen data.

A key factor in creating predictive and well-calibrated models using Dropout is the correct tuning of the Dropout probabilities for each layer of the neural network. This is typically done using a grid search over a number of Dropout values combined with a validation set to evaluate the posterior predictive distribution and the calibration of the obtained models \citep{gal2016dropout}. This process can be extremely time-consuming as for each parameter setting of the $p_i$ values in the grid search, as an entire network training and validation process has to be performed. For deep neural networks trained on large datasets, this can easily be on the order of days or weeks for individual networks, which can be prohibitive to practical application even with significant computational resources.

To alleviate these issues \citet{gal2017concrete} propose to use a distribution for binary discrete values that allows an application of the reparameterization trick so that one can optimize the values $p_i$ for each layer using back-propagation and gradient-based optimization methods. Using this approach, no grid search is necessary and all the Dropout probabilities are tuned for a given dataset during a single training process of the neural network. Instead of using the discrete Bernoulli distribution as part of the variational distribution (Equation~\ref{eq:var_bernoulli}), \citet{gal2017concrete} propose a continuous relaxation, the so-called Concrete distribution, which produces random variables in the continuous interval from $[0, 1]$ and concentrates its probability mass close to the bounds of this interval. The level of which the Concrete distribution concentrates its mass to the end-members is dependent on the temperature parameter $\tau$. Samples from the Concrete distribution can be obtained by transforming samples $u\sim\text{Uniform(0, 1)}$ with temperature $\tau$ and probability $p_i$
\begin{equation}
\begin{gathered}
    \displaystyle{\mathbf{z} = \sigma\left\{\frac{1}{\tau}\cdot(\log p_i - \log(1-p_i)+\log u - \log(1-u))\right\}},\\ 
    u\sim\text{Uniform}(0, 1)
\end{gathered}
\end{equation}

From the application of standard rules of calculus, we can also see that the partial derivative $\frac{\partial z}{\partial p}$ is independent of the noise variable $u$ and can be automatically obtained in an efficient manner through standard backpropagation frameworks \citep{paszke2019pytorch,tensorflow2015-whitepaper,jax2018github}.

To optimize the values $p_i$ for each layer in the neural network, an initial value $p_{init}$ has to be chosen, typically from the interval $p\in[0, 0.5]$. \citet{gal2017concrete} have shown that the optimization process of the individual layer Dropout probabilties for Concrete Dropout are robust to the initialization value $p_{init}$ and converge to the same values of $p_i$ independent of the initialization of $p_{init}$ and the weight matrices $\mathbf{W}_i$.

In our case, we extend the formulation of Concrete Dropout to CNNs as is given by \citep{gal2017concrete} and apply Concrete Dropout layers after every convolutional layer in our CNN. The initial Dropout probabilities $p_{init}$ for each Concrete Dropout layer were sampled from a uniform distribution \text{Uniform}(0, 0.5) and optimized during the training process of the CNN. The temperature $\tau$ was set to $2/3$ as recommended by \citet{gal2017concrete}. At test time we use $T=30$ samples from $q(\mathbf{\omega})$ on all synthetic benchmark evaluations and $T=10$ predictions for each seismic dataset.

\subsection{Stochastic Weight Averaging w. Gaussian Approximation (SWAG)}\label{sec:swag}
In many machine learning applications, we consider objective functions of the form 
\begin{equation}
    \mathcal{L}(\mathbf{\omega})=\sum_{i=0}^{N}l_i(\mathbf{\omega}, \mathbf{x}_{i}), \ \mathbf{g}(\mathbf{\omega}) := \nabla_{\mathbf{\omega}}\mathcal{L}(\mathbf{\omega})\label{eq:gradient}
\end{equation}
where $\mathbf{\omega}$ are the weights of our model, $l_i(\mathbf{\omega}, \mathbf{x}_{i})$ corresponds to an individual contribution to the loss from an input data $\mathbf{x}_i$, 
and $\mathbf{g}(\mathbf{\omega}, \mathbf{x}_{i})$ corresponds to the gradient of the objective function with respect to the model weights $\mathbf{\omega}$.

It has been shown that obtaining estimates of the full gradient $g(\cdot)$ in Equation~\ref{eq:gradient} using small batches of data, the so-called mini-batches, is beneficial when trying to find a solution to this non-convex optimization problem, and moreover because the memory requirements to perform full-gradient evaluation would be prohibitive of any practical application. This leads to a formulation of our loss with estimates of the gradient obtained over a mini-batch of data
\begin{equation}
    \hat{\mathcal{L}}_{B}(\mathbf{\omega})=\frac{1}{B}\sum_{n\in B}\hat{l}_n(\mathbf{\omega}, \mathbf{x}_{n}), \ \hat{\mathbf{g}}_{B}(\mathbf{\omega}) := \nabla_{\mathbf{\omega}}\hat{\mathcal{L}}_{B}(\mathbf{\omega})\label{eq:minibatch_loss}
\end{equation}

This objective function is then minimized using a first-order gradient-descent algorithm with mini-batch estimates of the full gradient 
\begin{subequations}
\begin{equation}
    \mathbf{\omega}(t+1) = \mathbf{\omega}(t)-\epsilon\hat{\mathbf{g}}_{B}(\mathbf{\omega}(t))
\end{equation}
\begin{equation}
\begin{gathered}
\hat{\mathbf{g}}_{B}(\mathbf{\omega}(t) = \mathbf{g}(\mathbf{\omega})+\frac{1}{\sqrt{B}}\Delta g(\mathbf{\omega}),\\
\Delta g(\mathbf{\omega})\sim\mathcal{N}(0, \mathbf{C}(\mathbf{\omega})) \end{gathered}\label{eq:gradient_noise}
\end{equation}
\end{subequations}
where $\epsilon$ is the step size, and $B$ indicates that the gradient is estimated over a mini-batch of data. We assume that the gradient noise follows a Gaussian distribution with covariance $\mathbf{C}(\mathbf{\omega})$, which is assumed time-independent and factorizes $\mathbf{C}(\mathbf{\omega})\approx \mathbf{C}=\mathbf{G}\mathbf{G}^T$ \citep{mandt2017stochastic}. This first-order method is equivalent to the well-known SGD algorithm. \citet{mandt2017stochastic} show that under certain assumptions, the SGD algorithm with mini-batch gradient estimates is equivalent to an Ornstein-Uhlenbeck process, a type of stochastic partial differential equation. The analytical stationary solutions of this stochastic process take on the form of Gaussian probability distributions
\begin{equation}
    q(\mathbf{\omega}) \propto \text{exp}\{-\frac{1}{2}\mathbf{\omega}^T\mathbf{\Sigma}^{-1}\mathbf{\omega}\}\label{eq:approxgauss}
\end{equation}
where the covariance $\Sigma$ satisfies the relationship
\begin{equation}
    \mathbf{\Sigma}\mathbf{A} + \mathbf{A}\mathbf{\Sigma} = \frac{\epsilon}{B}\mathbf{G}\mathbf{G}^T
\end{equation}
where $\mathbf{A}$ is the Hessian at the optimum value for $\omega$ that minimizes the loss $\hat{\mathcal{L}}_B(\mathbf{\omega})$.

From the relationship to the Ornstein-Uhlenbeck process and its solutions, \citet{mandt2017stochastic} derive an approach to sample from the full posterior distribution of the model weights given the training data $p(\mathbf{\omega}|\mathcal{D})$ by averaging SGD iterates using stochastic gradient estimates of the loss (Equation~\ref{eq:minibatch_loss}).

\citet{izmailov2018averaging} show that ensembling of the obtained weights $\mathbf{\omega}(t)$ from SGD iterates in the parameter space provides more robust weight configurations for deep neural networks which generalize more easily from training data $\mathcal{D}$ to unseen test data $\mathbf{x}^*$. Their stochastic weight averaging method (SWA) does not require the training of multiple models to obtain an ensemble of model outputs but can be obtained from a single training run with cyclical-learning rate schedules. Training is performed using a stochastic gradient descent method with a large step size to provide a burn-in phase that samples from different modes of the model weight distribution. After the burn-in phase, the learning rate is slowly annealed to a smaller step size and averaging of consecutive weight configurations of the model is initiated. During training, the weights of consecutive SGD iterations are averaged to form a stochastic weight average $\mathbf{\omega}_{SWA}$ which provides a robust configuration of model weights and at prediction time has been shown to generalize better than training a single model using standard stochastic gradient-descent only.

\citet{maddox2019simple} find approximate probabilistic representations of the SGD iterates obtained from SWA by representing the local basin of model weights $\mathbf{\omega}$ with a Gaussian distribution. This provides an approximate representation of the posterior distribution over model weights given training data along the lines of the solutions obtained by the analytical stationary distribution of the Ornstein-Uhlenbeck process (Equation~\ref{eq:approxgauss}).

First, a diagonal approximation of the covariance matrix is obtained by computing running averages of the SGD iterates of the model weights $\bar{\mathbf{\omega}}$

\begin{subequations}
\begin{equation}
    \bar{\mathbf{\omega}} = \frac{1}{T}\sum_{i=0}^{T} \mathbf{\omega}_{i}
\end{equation}
\begin{equation}
    \bar{\mathbf{\omega}^2} = \frac{1}{T}\sum_{i=0}^{T} \mathbf{\omega}^2_{i}
\end{equation}
\begin{equation}
        \mathbf{\Sigma}_{diag} = \text{diag}(\bar{\mathbf{\omega}^2}-\bar{\mathbf{\omega}}^2)
\end{equation}
\begin{equation}
    \mathbf{\omega}\sim \mathcal{N}(\bar{\mathbf{\omega}}, \mathbf{\Sigma}_{diag})\label{eq:diag}
\end{equation}
\end{subequations}

In the case of deep neural networks, the diagonal approximation can be overly simplistic \citep{blier2018desc} therefore \citet{maddox2019simple} approximate the full-rank covariance $\mathbf{\Sigma}$ with a low-rank approximation
\begin{equation}
    \mathbf{\Sigma} \approx \frac{1}{T-1}\sum_{i=0}^{T} (\mathbf{\omega}_i-\bar{\mathbf{\omega}})(\mathbf{\omega}_i-\bar{\mathbf{\omega}})^T = \frac{1}{T-1}DD^T
\end{equation}
of rank $T$, where the columns of $D$ are the deviation from the mean $(\mathbf{\omega}_i-\bar{\mathbf{\omega}})$. Furthermore, a full rank representation for neural networks would be infeasible for deep neural networks, therefore a low-rank approximation of rank $K$ can be obtained by computing the deviation matrix $\hat{D}$ where only the last $K$ SGD iterates are used to construct the covariance matrix $\Sigma_{low}$
\begin{equation}
\mathbf{\Sigma}_{low} = \frac{1}{K-1}\hat{D}\hat{D}^T
\end{equation}
and finally, we obtain samples from the approximate posterior distribution of the weights by combining the diagonal and low-rank approximation of the covariance 
\begin{equation}
    \widetilde{\mathbf{\omega}} \sim \mathcal{N}\left( \bar{\mathbf{\omega}}, \frac{1}{2}(\mathbf{\Sigma}_{diag}+\mathbf{\Sigma}_{low})\right)\label{eq:lowrank}
\end{equation}

In practice, we obtain samples from the Gaussian approximation for models with a total number of weights $d$ 
\begin{subequations}
\begin{equation}
    \widetilde{\mathbf{\omega}}= \bar{\mathbf{\omega}}+\frac{1}{\sqrt{2}} \mathbf{\Sigma}_{diag}^{\frac{1}{2}}z_1+\frac{1}{\sqrt{2(K-1)}}\hat{D}z_2
\end{equation}\label{eq:sampleswa}
\begin{equation}
    z_1 \sim \mathcal{N}(0, I_d) \ \text{and} \ z_2 \sim \mathcal{N}(0, I_K)
\end{equation}
\end{subequations}

At inference time, when making predictions on unseen test data, we obtain samples from the weight posterior (Equation~\ref{eq:bayes}) and characterize the predictive posterior distribution as outlined previously (Equation~\ref{eq:montecarlomean}), as well as compute uncertainties (Equation~\ref{eq:mutual_information}) from the predictions of the individual weight realizations (Equation~\ref{eq:sampleswa}).

We closely follow the training procedure and learning rate schedule proposed by \citet{maddox2019simple} and \citet{wilson2020bayesian}. We use a constant learning rate of $\text{lr}_{SWA}=10^{-2}$ and linearly anneal the learning rate to the sampling learning rate $\text{lr}_{SWA}=10^{-3}$ from where we obtain the SGD parameter iterates to construct the diagonal (Equation~\ref{eq:diag}) and low-rank approximations (Equation~\ref{eq:lowrank}) of the Gaussian posterior approximation. 

\subsection{Additional Baselines}\label{sec:additional_baselines}
In addition to the probabilistic methods presented (see also Table~\ref{tab:methods}), we provide two additional baseline methods as they represent typical approaches that are applied in geophysical applications such as fault detection. We train two additional networks, one with an unbalanced BCE-loss and $L_2$ weight regularization (Unbalanced-CE), and a second with a balanced BCE-loss as proposed by \citep{wu2019faultseg3d} (Balanced-CE). Using a weighted BCE-loss is a popular practical choice when dealing with imbalanced data.

It is worth mentioning that training a single neural network with a MAP objective function as is the case for these two additional baselines, Unbalanced- and Balanced-CE,  means that marginalization over the posterior weight distribution is not possible as we only consider a single weight realization and hence both these approaches do not provide reliable uncertainty estimates.

\section{Datasets and Evaluation Metrics}\label{sec:datasets}
\subsection{Training and Test Data Generation} \label{sec:data_generation}
\subsubsection{Synthetic Datasets}\label{sec:synthetic_datasets}
\begin{figure*}
    \centering
    \includegraphics[width=\textwidth]{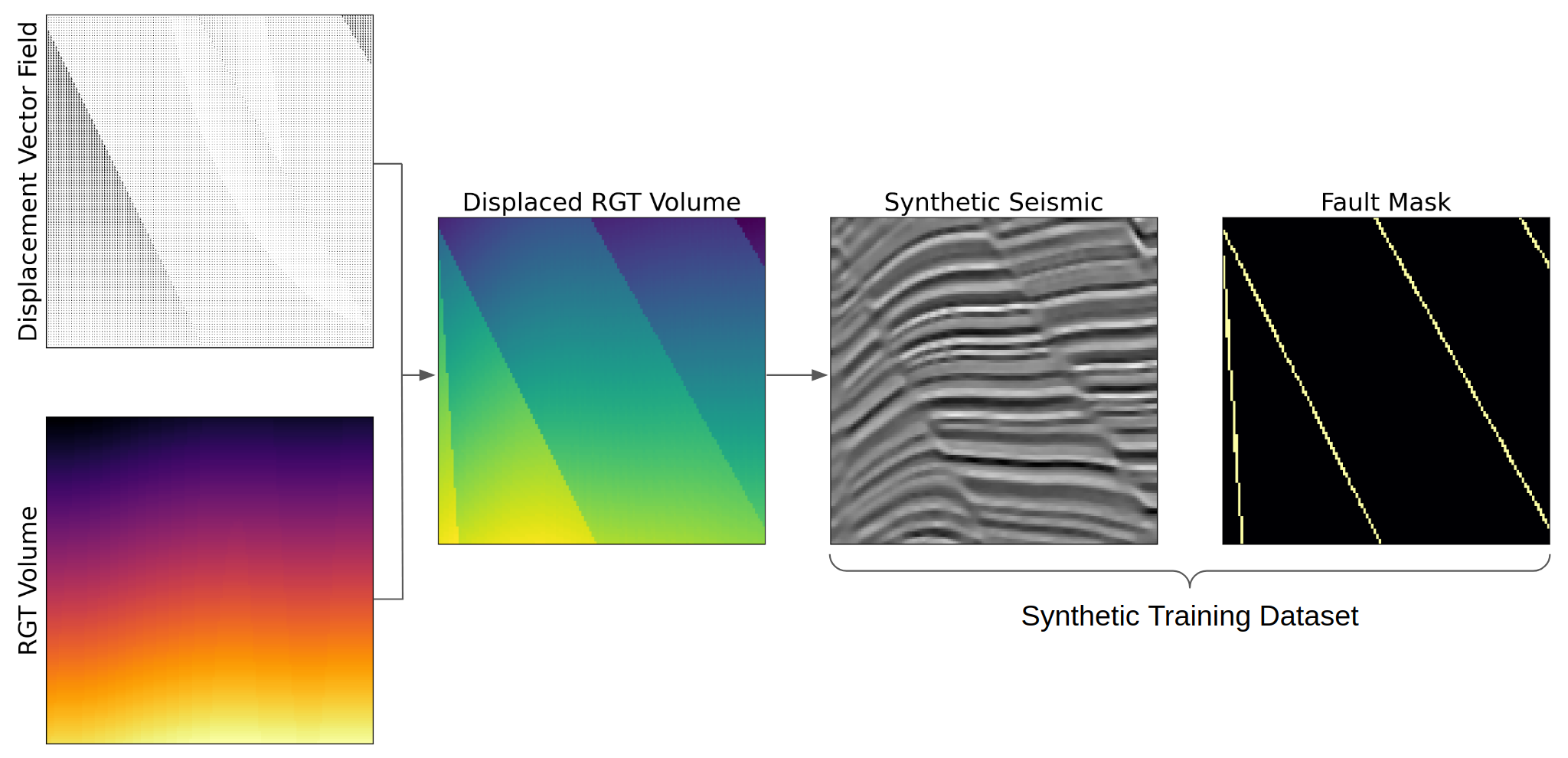}
    \vspace{2pt}
    \caption{Generation of synthetic seismic data based on a penny-shaped fault displacement model and a stochastic seismic forward modeling process.}
    \label{fig:faults}   
\end{figure*}
Obtaining labels of geological faults within real seismic datasets is a time-consuming manual task with a high degree of uncertainty. Precise placement of fault labels based on image data is difficult due to resolution issues close to fault planes, seismic processing and imaging artifacts, and a lack of possible calibration data away from boreholes. 

One approach to overcome this is to generate high-quality fault label sets using synthetic seismic images as proposed by \citet{wu2019building}. To generate seismic faults we assume a penny-shaped displacement model for the fault displacement, where displacement at the fault-tips is zero and maximal at the center of the fault plane. In our dataset, we only consider normal faults, while we note that the synthetic data generation process could be generalized to other faulting regimes such as reverse or strike-slip faulting.

To create a large number of unique fault configurations and seismic scenarios we consider the synthetic data generation process as a data augmentation problem.
Furthermore, due to the volumetric nature of the seismic images, large datasets of seismic image patches can quickly require large amounts of storage when generated prior to training. To minimize the storage requirements necessary in the training process and to maximize the diversity of the generated images, we precompute $N=4000$ fault displacement fields prior to starting the network training process for each of the methods presented in Table~\ref{tab:methods}. At training time and for each mini-batch of $k$ training images, we generate $k$ unique Gaussian random fields (GRF) for each fault patch which are used to create an unfaulted relative geological time volume (RGT), based on the method presented by \citet{wu2019building}. The unfaulted RGT volumes are then displaced by their corresponding fault displacement vector fields to generate a faulted RGT volume (Figure~\ref{fig:faults}).
These RGT volumes are then randomly populated with reflection coefficients that are then subsequently forward modeled using a 1D convolution approximation. Ricker and Ormsby wavelets are generated with randomly sampled wavelet frequencies and used to create synthetic seismic images.

By applying dip-steered edge-preserving smoothing of the generated synthetic we reduce the presence of aliasing artifacts in the generated synthetic images. We then add random Gaussian and structured Gaussian noise to the seismic images leading to synthetic seismic images of varying fidelity and noise characteristics.

Real seismic data often have areas where no faulting is present within a seismic patch, we therefore also randomly generate synthetic seismic images with no faults present within the training example as to minimize the possibility for the network to generate false positives.

The entire process outlined here to generate synthetic seismic images can be performed on the fly at training time. While we only generate 4000 fault vector displacement fields, every batch element that is used to train the networks has exact fault labels and represents a unique seismic expression with randomly generated geological background models and imaging artifacts. This allows us to increase the diversity of the generated dataset by orders of magnitude compared to previous approaches \citep{wu2019building, wu2019faultnet}.

To monitor the training process of our networks, we split our dataset into a training and validation dataset with 3600 and 400 fault displacement fields respectively. The synthetic images used for validation are forward-modeled and remain the same during the entire training process. Finally, we have generated a test set of 400 pre-computed synthetic fault images and label masks which we use to quantitatively evaluate the considered methods.

\subsubsection{Real Seismic Datasets}\label{sec:real_datasets}
We present results on three seismic datasets with various faulting regimes where each dataset also has specific noise and imaging characteristics. First, we consider a public seismic dataset from offshore Australia located in the Canning basin \citep{yule2019canning} which has undergone multiple extensional and rifting events and is highly faulted. The dataset represents a subset of a larger PSTM volume covering a total area of 4446 $\text{km}^2$.
The second dataset is known as F3 that was acquired in the Dutch North-Sea and has become a standard dataset for evaluating the capabilities of various fault detection algorithms \citep{wu2019faultseg3d}. While there is one major long fault that extends from the salt-diapir into the shallow parts of the geology, the set of faults commonly used to compare the obtained fault probability attributes is primarily present in the deeper section of this seismic data where migration artifacts and noise can dominate in places. Finally, a dataset from the Norwegian North-Sea (SUN12), with high-quality imaging of faults is presented. The section shown here represents a subset from a shallow polygonal faulted layer.

\subsection{Neural Network Architecture and Training}\label{sec:architecture}
The architecture of a neural network, i.e. the arrangement and sequence of the various layers and activation functions that define a neural network, can have a major impact on the performance of a Deep Learning-based algorithm for a specific task. As outlined previously, we treat the task of fault-detection within a seismic dataset as a pixel-wise semantic segmentation problem.

For each three-dimensional seismic image, we seek to predict for every seismic sample whether there is a fault present at the specific location or not. This is a common task in Deep Learning applications for computer vision across many sciences and industries such as biomedical imaging.

The most successful architectures for semantic segmentation consist of two distinct parts: the first section of the network applies a sequence of convolutional layers followed by an operation that reduces the spatial dimensions of the input images and the intermediate features. This downsampling operation can be achieved through the use of so-called Max-Pooling or Average-Pooling, as well as a strided convolution operation. Typically, while the spatial dimensions of the image decrease throughout the first part of the network, the number of features produced by each consecutive convolutional layer, is increased. The final layer of this contracting part of the network consists of a representation with small spatial dimensions e.g. $16^3$ samples but hundreds of feature maps. The guiding principle is that this high-dimensional representation encodes high-level and large-scale attributes of the high-resolution input images, hence this first part of the network is often called the encoder. The task of the second part of the networks is to create high-resolution predictions at the scale of the original input images, hence the second part is called the decoder. The decoder achieves this by a sequence of convolutional layers followed by an upsampling operation such as a nearest-neighbor, a trilinear interpolation, or a so-called transposed convolution. The final layer of the decoder represents the pixel-wise logits used to make the final prediction for each of the pixel-wise semantic labels present in the dataset. \citet{ronneberger2015u} introduced additional connections between the encoder and decoder of the network architecture which concatenate or add feature maps of intermediate layers of the encoder, to feature maps of the decoder with the same spatial resolution. Their so-called U-Net architecture has become state-of-the-art for many semantic segmentation tasks including seismic fault detection \citep{wu2019faultseg3d}.

\begin{figure*}
    \centering
    \subfloat[Schematic of the CNN architecture used in this study, and color-legend of the individual operations of the convolutional blocks of the network.]{
        \includegraphics[width=\textwidth]{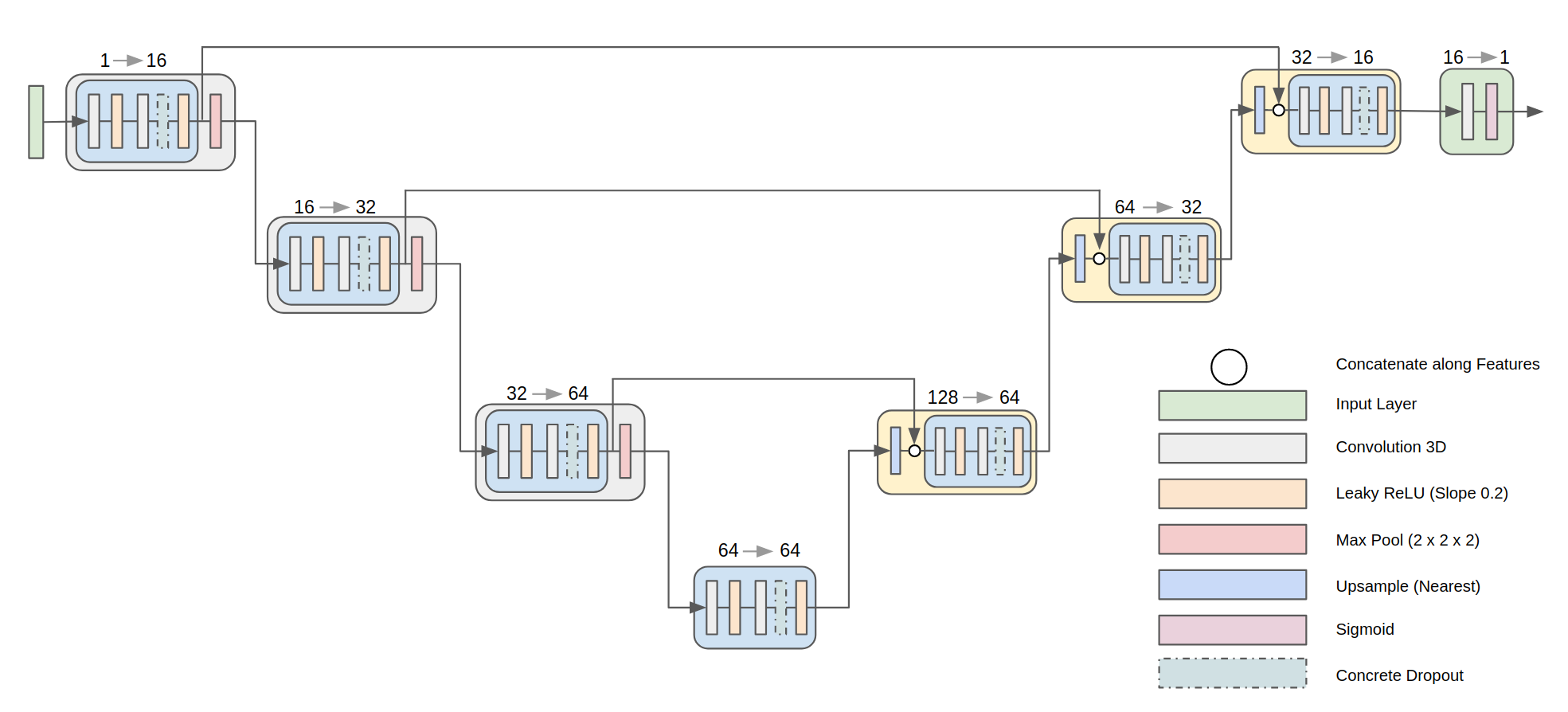}
        \label{fig:convblocks}   
        } \\
    \subfloat[Schematics of the convolutional blocks used to build up the architecture of the CNN used. Dropout layers are inserted into the double-convolution block only in the case of Concrete Dropout.]{
        \includegraphics[width=\textwidth]{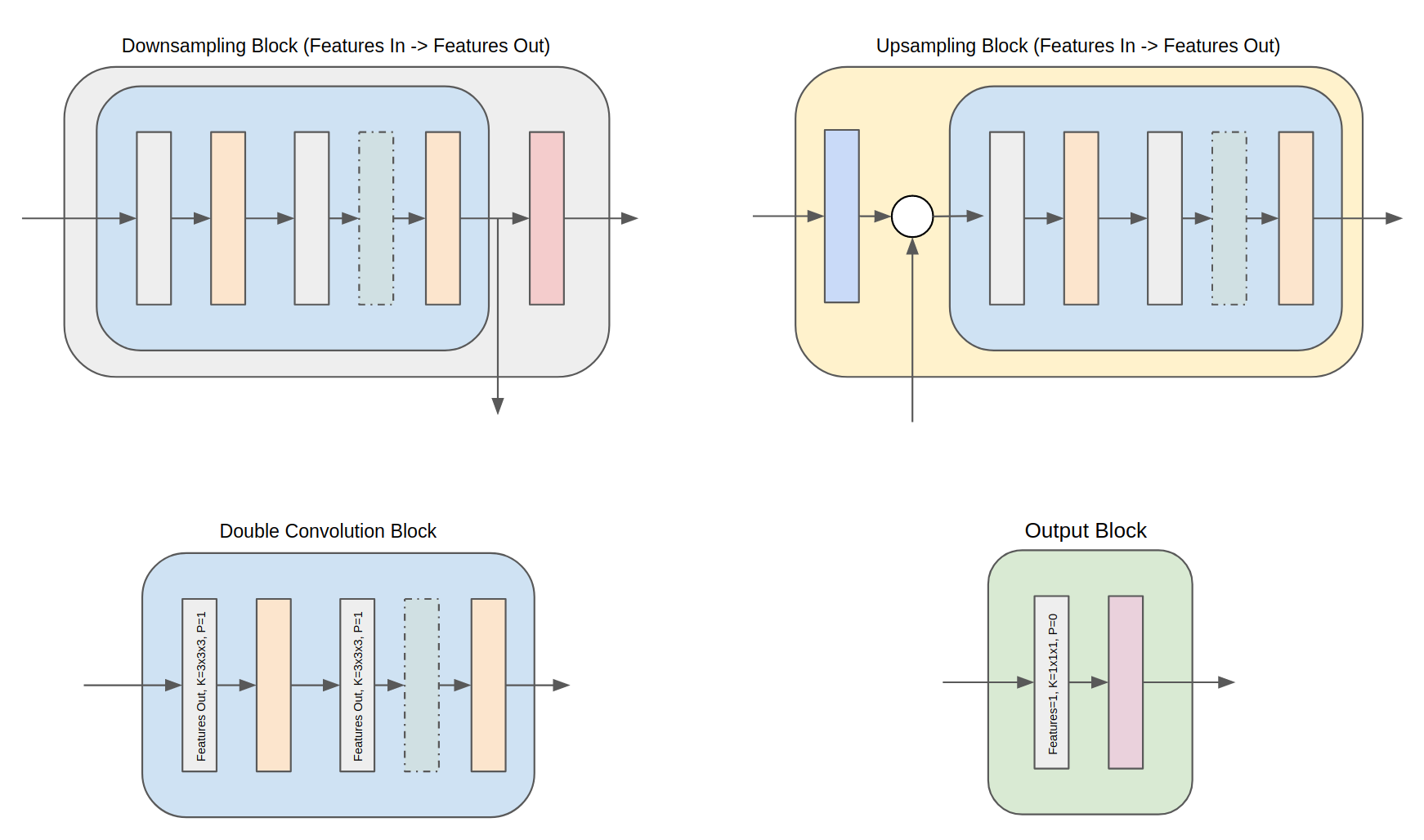}
        \label{fig:convnet}
    }
    \caption{Neural network architecture and configuration of individual network block configurations. The model closely follows the U-Net \citep{ronneberger2015u} architecture.}
\end{figure*}

In our case, we use a U-Net architecture that closely resembles the network proposed by \citet{wu2019faultseg3d}. The network is organized into so-called blocks which represent a grouped series of convolutional layers and activation functions. Each convolutional block consists of two convolutional layers followed by a Leaky-ReLU activation function. The convolutional operators in all double convolutional blocks consist of three-dimensional convolutional layers with $3\times3\times3$ kernels, a stride of 1, and 1 sample of zero-padding applied to all sides of the input to the convolutional layer. The down-sampling block consists of a double convolution block shown in Figure~\ref{fig:convblocks} followed by a $2\times2\times2$ Max-Pool operator, where the feature maps prior to the Max-Pool operator are passed via the skip-connection to the upsampling blocks of the decoder. These upsampling blocks, first apply a trilinear interpolation to the input features from the previous upsampling block and are concatenated with the feature maps received from the encoder via the skip-connection. In total the encoder consists of 3 downsampling blocks with an additional double-convolutional block, which is followed by 3 upsampling blocks of the encoder as shown in Figure~\ref{fig:convnet}. The final predictions are made by an output block that consists of a single $1\times1\times1$ convolution operation and a sigmoid activation. The sigmoid activation projects the logits into a continuous interval between 0 and 1, where a value close to one for a specific sample, represents the presence of a fault at that location. For all methods presented in Table~\ref{tab:methods} we use the network architecture presented in Figure~\ref{fig:convblocks} where for the case of the Concrete Dropout method, we introduce a Concrete Dropout-layer after the second convolution operation in every double convolutional block.

\subsection{Evaluation Metrics}\label{sec:metrics}
One of the objectives of this study is to perform a quantitative and qualitative analysis of the different methods used to obtain probabilistic predictive models for fault segmentation. To assess the performance quantitatively, however, one has to evaluate some metrics that capture the ability of each network to produce a calibrated fault probability attribute.

The Negative-Log-Likelihood (NLL) is a proper scoring rule that heavily penalizes overconfident predictions. Therefore, the NLL also depends on the prediction uncertainty of our models and has become a popular and simple measure to evaluate and compare the calibration of probabilistic models. 

In the case of geological fault segmentation, we compute the average NLL over the $N$ members of the test dataset for each sample in the synthetic seismic images
\begin{equation}
    \mathcal{L}_{NLL} = -\frac{1}{N}\sum_i^N y_i\log(\hat{p}_i)-(1-y_i)\log(1-\hat{p}_i)\}
\end{equation}
where $\hat{p}_i$ is the predicted probability (Equation~\ref{eq:probabilistic_model}). While the NLL is useful as a metric to measure calibration, it is typically compromised by other regularisation strategies, and maximizing performance in terms of other metrics such as accuracy or IoU can be achieved at the cost of poor calibration \citep{guo2017calibration}. 

For binary classification tasks such as the pixel-wise fault segmentation problem considered here, the Brier score \citep{gneiting2007probabilistic}, which is defined as the mean-squared error between the assigned label and the assigned probability
\begin{equation}
    \mathcal{L}_{Brier} = \frac{1}{N}\sum_{i}^{N}\left(y_i-\hat{p}_i\right)^2
\end{equation}
is also a proper scoring rule which we compute as an average over all image pixels of the test set of synthetic seismic images.

Finally, we use the expected calibration error (ECE) proposed by \citep{naeini2015obtaining} as an additional measure of miscalibration. The ECE is derived as a measure for the expectation in difference between accuracy and confidence
\begin{equation}
\mathop{\mathbb{E}}_{\hat{P}}\left[\left|\mathop{\mathbb{P}}\left(\hat{Y}=Y|\hat{P}=p\right)-p\right|\right]
\end{equation}
which we discretize into equally-sized bins and compute a weighted average based on the number of samples within each probability-bin
\begin{subequations}
\begin{equation}
    ECE = \sum_{m=1}^M \frac{|B_m|}{n}\left|acc(B_m)-conf(B_m)\right|
\end{equation}
\begin{equation}
    acc(B_m) = \frac{1}{|B_m|}\sum_{i\in B_m} \mathbf{1}(\hat{y}_i=y_i)
\end{equation}
\begin{equation}
    conf(B_m) = \frac{1}{|B_m|}\sum_{i\in B_m} \hat{p}_i
\end{equation}
\end{subequations}
where $n$ is the total number of pixels in the test set, $|B_m|$ is the number of pixels $n$ in the $m$-th bin, $\hat{y}_i$ and $y_i$ are the label and predicted label respectively, $\mathbf{1}(\cdot)$ is the indicator function \citep{guo2017calibration}. We use $M=15$ bins to compute the ECE for the test set of synthetic seismic images and their fault labels.

\section{Results}\label{sec:results}
\subsection{Model Training}\label{sec:training}
\begin{figure*}
    \centering
    \includegraphics[width=\textwidth]{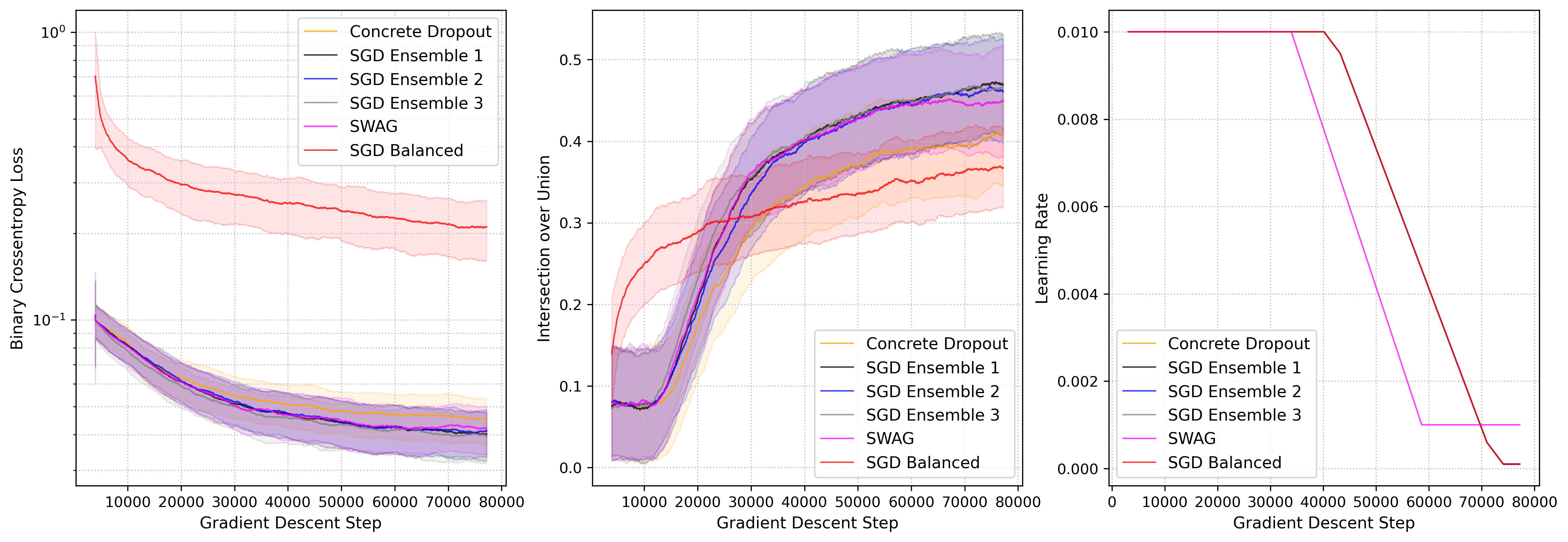}
    \caption{Training loss, IoU, and learning rate schedule as a function of the training iterations for each of the proposed methods.}
    \label{fig:learning}   
\end{figure*}

All networks were trained with a total of 77000 gradient descent updates. We note that while the concept of epochs i.e. a full pass over a dataset is commonly used throughout the machine learning literature, an epoch can consist of a variable number of gradient-based updates to the weights of the neural network when using different sizes of mini-batches. While the mini-batch size of 16 has been selected for all methods, we do believe that this provides a more objective representation of the optimization process, or stochastic sampling process, in the case of SWAG. During training nonetheless, we monitor two key metrics, the BCE-loss, and the IoU computed as an average over the mini-batch elements for the training set, and as an average evaluated over the validation set. In Figure~\ref{fig:learning} we show the mean (solid lines) and standard deviation (shaded colored area) of the BCE loss and IoU computed as a running average over 1000 gradient update steps. 

Each individual CNN was initialized with a different set of initial weights. The learning rate (Figure~\ref{fig:learning}, right) was linear annealed from $10^{-2}$ to $10^{-4}$ for all methods except SWAG, which requires a higher learning rate during the sampling process (Equation~\ref{eq:approxgauss}). The model corresponding to training with SGD and an unbalanced binary cross-entropy corresponds to Deep Ensemble member 1. All models were trained on 4 Nvidia V100 GPUs where each training run required approximately two days of training to perform 77000 gradient updates, the total run-time for the final training process of the models shown here amounts to roughly two weeks of training time. Training was performed in parallel for each method, and all networks and probabilistic methods were implemented in the automatic differentiation framework PyTorch \citep{paszke2019pytorch}.

We can see distinct differences between the different approaches based solely on the NLL and the IoU (Figure~\ref{fig:learning}, left and center). Most prominently we can see that the balanced loss for the Balanced-CE model has a much higher numerical value, due to the applied weighting, and moreover shows the lowest IoU at the end of training. A key difference that can be observed is that the balanced loss function leads to a rapid increase in the IoU during the initial training process. This can be beneficial when a network needs to be trained within a few iterations or when only a small training budget is available. Contrary to the balanced loss, all other methods, which do not use any balancing of the fault and background labels, are not able to detect any faults initially and require approximately 25000 SGD steps to reach an IoU equivalent to the balanced loss, and quickly exceeding the balanced approach in measured IoU thereafter.

For all methods, we observe a high variance of the BCE and IoU shown by the broad $\pm\sigma$ region around the running mean for each metric. This can be explained by the high diversity of the synthetic training dataset generated during the training process, and may additionally hint at the need for larger batch sizes during training. In our case study, due to computational constraints with regards to available memory of the graphics processing units, training with larger batch sizes was not feasible, however, it can be considered as future work to further improve the presented results. Furthermore, it can be seen based on Figure~\ref{fig:learning} that all methods show a positive trend that with an even higher number of iterations, the model performance could still be further improved.

\subsection{Evaluation on Synthetic Datasets}\label{sec:evaluation_synthetic}
Evaluating the proposed methods and baselines on a synthetic dataset has the advantage to allow us to perform quantitative and qualitative comparisons in terms of model calibration and uncertainty quantification. As mentioned before, based on the test dataset of 400 pre-computed synthetic seismic volumes we evaluate the metrics as an average over the entire test dataset. 

Table~\ref{tab:metrics} shows a comparison of the performance metrics for calibration and segmentation of each of the considered methods. The model trained with an unbalanced CE-loss shows the lowest NLL, and second-best Brier-Score, ECE, and IoU amongst all the methods compared. This indicates that a well-calibrated predictive model can be obtained using an unbalanced loss function. This is in stark contrast to the model trained using a balanced CE-loss, which shows poor calibration based on the evaluated metrics. Based, on the Brier-Score and the ECE, we see that the Deep Ensemble outperforms not only the other probabilistic methods but also the deterministic unbalanced CE baseline. This reflects the common notion that Deep Ensembles are a good baseline for probabilistic models, which come at a high computational cost. Finally, SWAG outperforms Concrete Dropout in all calibration metrics, with equal predictive performance based on the IoU score. 

\begin{table*}
\centering
\ra{1.3}
\begin{tabularx}{\textwidth}{XXXXX}
\toprule
	Method & NLL $\downarrow$ & Brier Score $\downarrow$ & ECE $\downarrow$ & IoU $\uparrow$ \\
	\midrule
	Unbalanced-CE & $\mathbf{2.53e-03}$ & 3.35e-02 & 9.19e-03 & 0.52 \\
	Balanced-CE & 1.89e-02 & 1.02e-01 & 3.00e-02 & 0.40 \\
	Deep Ensemble & 2.85e-03 & $\mathbf{3.28e-02}$ & $\mathbf{9.01e-03}$ & $\mathbf{0.55}$ \\
	Concrete Dropout & 3.73e-03 & 3.65e-02 & 9.95e-03 & 0.51 \\
	SWAG & 2.74e-03 & 3.53e-02 & 9.66e-03 & 0.51 \\
	\bottomrule
	\end{tabularx}
\caption{Quantitative model performance for calibration and predictive accuracy on a test dataset of synthetic seismic images (N=400). Arrows indicate whether higher or lower values reflect a better metric value. Best scores are highlighted in bold.}\label{tab:metrics}
\end{table*}

\begin{figure}
    \centering
    \includegraphics[width=\columnwidth]{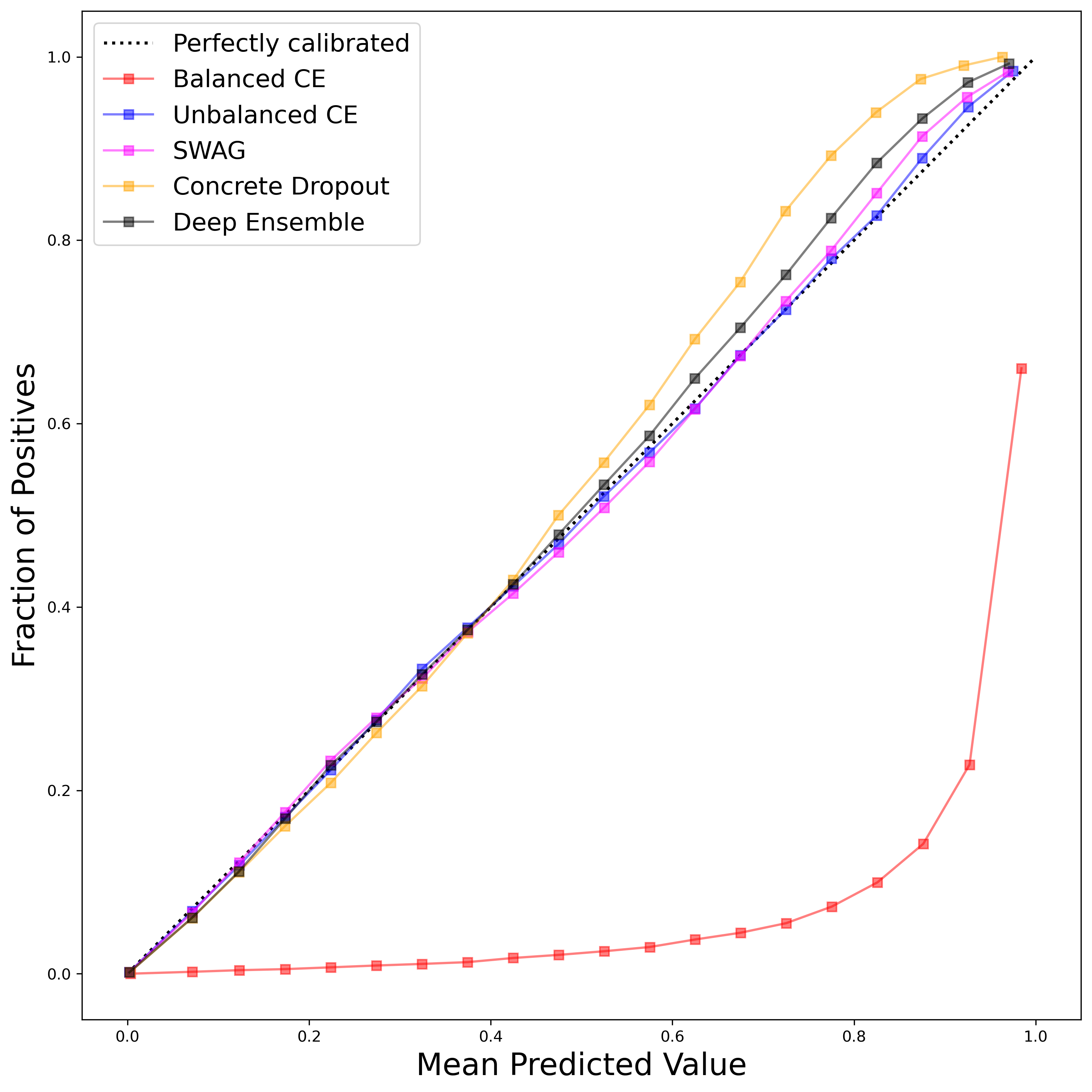}
    \caption{Comparison of the calibration curves for the proposed deterministic and probabilistic methods}
    \label{fig:calibration_curves}   
\end{figure}

The results presented in Table~\ref{tab:metrics} are further reinforced by the calibration plot shown in Figure~\ref{fig:calibration_curves}. We observe that the balanced CE loss leads to a significant overestimation of the fault probabilities leading to significant miscalibration. All other methods show good calibration throughout the entire range of predicted values, with minor underestimation of the true probabilities at the upper end of the probability range in the case of Concrete Dropout. 

\begin{figure*}
    \centering
    \includegraphics[width=\textwidth]{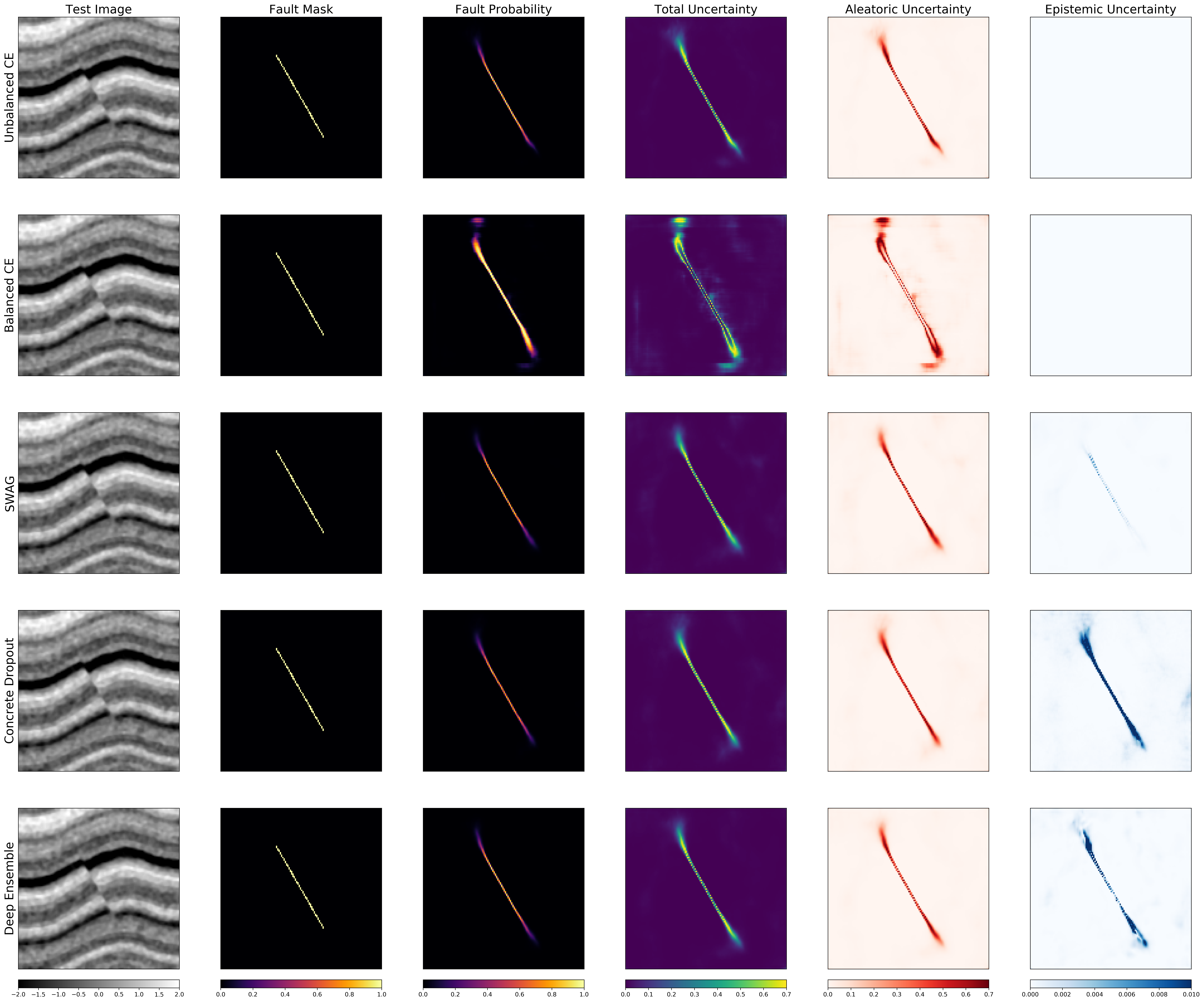}
    \caption{Comparison of the predicted fault probability and associated total, aleatoric, and epistemic uncertainties for penny-shaped normal fault in a 3D synthetic seismic volume. }
    \label{fig:synthetic_mid_uncertainties}   
\end{figure*}

Based on the examples from the synthetic test dataset where the location of the fault plane is known precisely, we can visualize the predicted pixel-wise fault probabilities, total, aleatoric, and epistemic uncertainties (Equation~\ref{eq:mutual_information}) for a two-dimensional section of a three-dimensional seismic dataset. In Figure~\ref{fig:synthetic_mid_uncertainties} we present a synthetic seismic image with a single penny-shaped normal fault with moderate displacement. We use this example to highlight differences in the uncertainties computed from the predictions obtained by each of the proposed deterministic and probabilistic methods.

For the deterministic methods, using the unbalanced and balanced CE-loss (Figure~\ref{fig:synthetic_mid_uncertainties}-row 1 and 2), we see that these approaches do not capture any epistemic uncertainty. While we can numerically evaluate the total uncertainty, this estimate is unreliable as it is based only on a single mode from the posterior weight distribution. Furthermore, we observe high uncertainties for all methods at the tips of the fault, and in the lateral placement of the fault plane. The fault probability map of the model trained with the balanced loss (Figure~\ref{fig:synthetic_mid_uncertainties}-row 2) shows a high fault probability where no fault is present leading to a high degree of false positives and a low IoU score. All three probabilistic methods (Figure~\ref{fig:synthetic_mid_uncertainties}-rows 3 to 5) show similar probabilities, total, and aleatoric uncertainties. The epistemic uncertainties of the three probabilistic methods are qualitatively different and a much smaller uncertainty can be observed for SWAG than for the case of Concrete Dropout and the Deep Ensemble.

\begin{table*}
\centering
\ra{1.3}
\begin{tabularx}{\textwidth}{XXXXX}
\toprule
			Method & NLL $\downarrow$ & Brier Score $\downarrow$ & ECE $\downarrow$ & IoU $\uparrow$ \\
			\midrule
			Unbalanced-CE & 8.34e-03 & 6.73e-02 & 1.51e-02 & 0.24 \\
			Balanced-CE & 3.41e-02 & 2.15e-01 & 6.42e-02 & 0.18 \\
			Deep Ensemble & $\mathbf{8.13e-03}$ & $\mathbf{6.49e-02}$ & $\mathbf{1.48e-02}$ & $\mathbf{0.33}$ \\
			Concrete Dropout & 9.48e-03 & 6.95e-02 & 1.55e-02 & 0.29 \\
			SWAG & 9.02e-03 & 7.29e-02 & 1.55e-02 & 0.26 \\
		\bottomrule
		\end{tabularx}
	\caption{Model performance on test dataset of synthetic seismic images with 20\% salt-and-pepper noise (N=400).}\label{tab:saltpepper5}
\end{table*}

A basic assumption when creating any probabilistic model is that the distribution of the training data should be the same as the distribution of data that the model will be applied on.
In many applications, this assumption does not hold since it may either be difficult to acquire such a training dataset, or there may be unforeseen changes in the distribution of the data where a model is to be applied. In the case of 3D seismic data, acquiring a training dataset based on actual 3D seismic field data with labeled faults would represent a time-consuming and expensive task, that also comes with a high degree of bias due to the lack of ground truth information on the correct identification of faults in seismic data \citep{bond2007you}. To overcome this, synthetic data can be created where the precise location of the faults is known. By relying on synthetic training data, the assumption is made that the synthetic training data distribution reflects the true data distribution encountered in 3D seismic data, and hence the performance evaluated on a synthetic test set should reflect the performance of the model on real seismic data. Nevertheless, synthetic data may actually only cover a subset of the geological faulting, seismic noise characteristic, and artifacts encountered in 3D seismic data. Therefore synthetic test datasets have an optimistic bias with respect to the performance of these models when applied to real seismic data. 

To evaluate the robustness of the methods considered towards a shift in the data distribution of the test dataset, we first evaluate the models obtained from each approach on the same test dataset with added salt-and-pepper noise (modeled by replacing a portion of seismic samples at random by the maximum amplitude of the given seismic volume). This represents a type of noise that can occur in seismic datasets, which the models were not trained on. 

Table~\ref{tab:saltpepper5} shows the result of the quantitative evaluation for the calibration and predictive metrics as an average over the synthetic test dataset when 20\% salt-and-pepper noise was added on a pixel-wise basis to the images. We see that the Deep Ensemble provides the highest predictive performance in terms of IoU, as well as the best calibration scores when a small artificial train-test shift is introduced. Hence the Deep Ensemble represents the most robust method against data distribution shift based on the evaluated metrics. Of the approximate Bayesian methods, Concrete Dropout slightly outperforms SWAG in the Brier Score and in terms of IoU. The balanced-CE approach also shows strong miscalibration for out-of-distribution data in this case, and unbalanced-CE shows good performance also in the case of this specific out-of-distribution noise.

\begin{figure*}
    \centering
    \includegraphics[width=\textwidth]{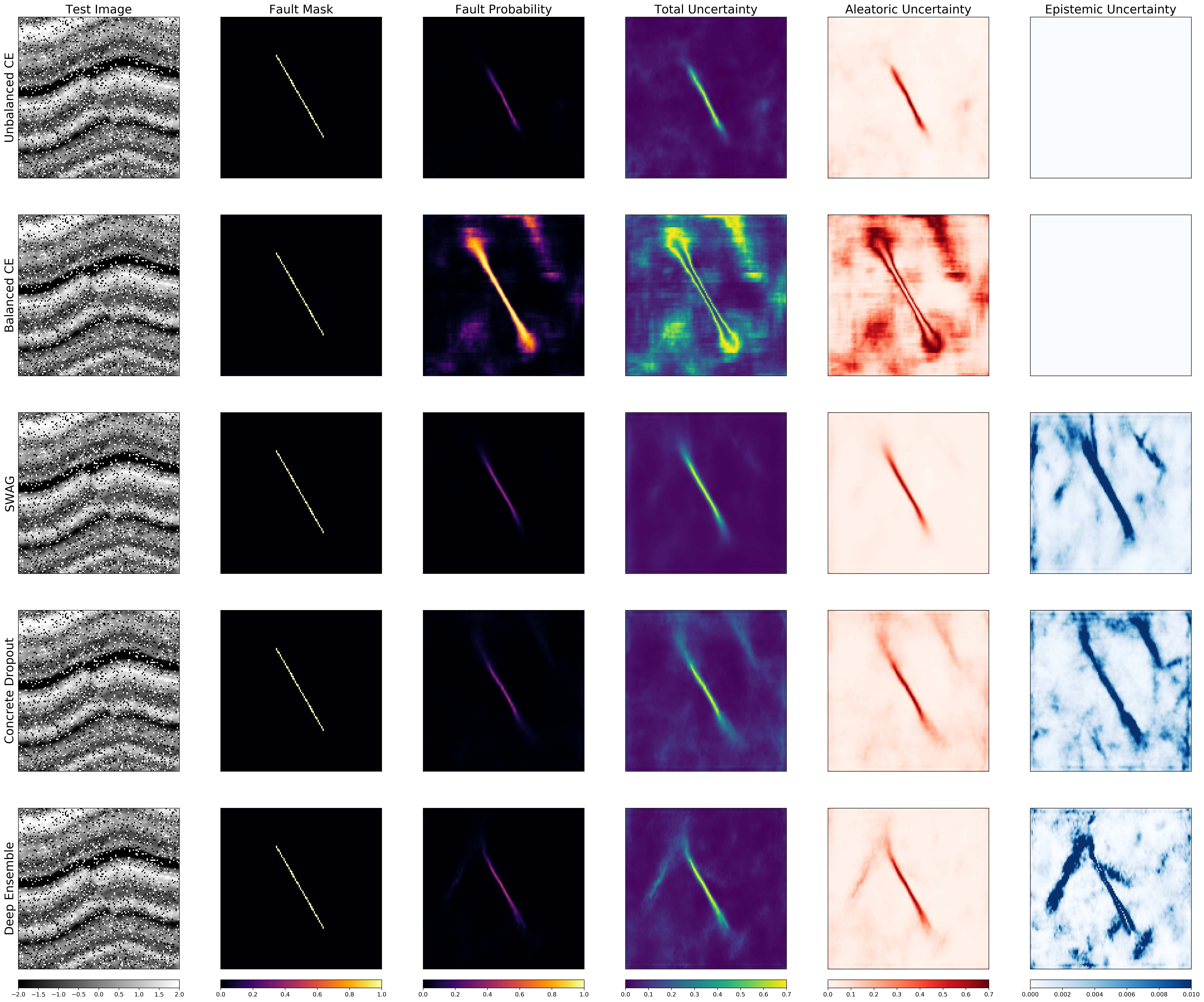}
    \caption{Comparison of the predicted fault probability and associated total, aleatoric, and epistemic uncertainties for penny-shaped normal fault in a 3D synthetic seismic volume with 20\% salt-and-pepper noise added.}
    \label{fig:synthetic_mid_uncertainties_noise}   
\end{figure*}

When evaluating the obtained pixel-wise probabilities and uncertainties with the added salt-and-pepper noise, we observe that for all probabilistic methods, the epistemic uncertainty considerably increases. This is expected since the models were not exposed to this type of noise artifact in the training process. Including this type of noise into the data-augmentation during training would considerably reduce the sensitivity of these models to salt-and-pepper noise and therefore reduce the epistemic uncertainty for synthetic images with these artifacts. Compared to the network trained with the unbalanced CE loss, the balanced loss induces false positives away from the actual fault plane, which cannot be observed in any of the other examples, and therefore, predictions that are less robust to a shift in the data distribution caused by the additional noise artifacts. Qualitatively we observe that for all three probabilistic methods, the total and in parts the aleatoric uncertainty is increased in areas with strong noise artifacts. The effect can be seen least on the SWAG model, and strongest on the Deep Ensemble example. 

To provide some insight as to why these noise artifacts can also lead to higher aleatoric uncertainty, we have to recall the definition of the aleatoric uncertainty itself. Generally speaking, aleatoric uncertainty is independent of the model weights which means predictions made by each realization of the weights from the posterior given the training data lead to some indication of a fault being present at these areas of high aleatoric uncertainty. Therefore, strong correlations in the noise, are commonly mistaken as faults in the seismic volume. Furthermore, it's important to note, that all methods are approximate and hence also the estimated uncertainties are only approximations, which itself are a result of a training process on a limited set of synthetic training images. One can conclude that these artifacts in the aleatoric uncertainty warrant further investigation on the quality of the uncertainty attributes themselves, and the compounding effects of the training dataset, and out-of-distribution data shifts. 
\subsection{Application to Seismic Datasets}\label{sec:evaluation_real}
\begin{figure*}
\centering
\subfloat[\centering Deep Ensemble]{{\includegraphics[width=5cm]{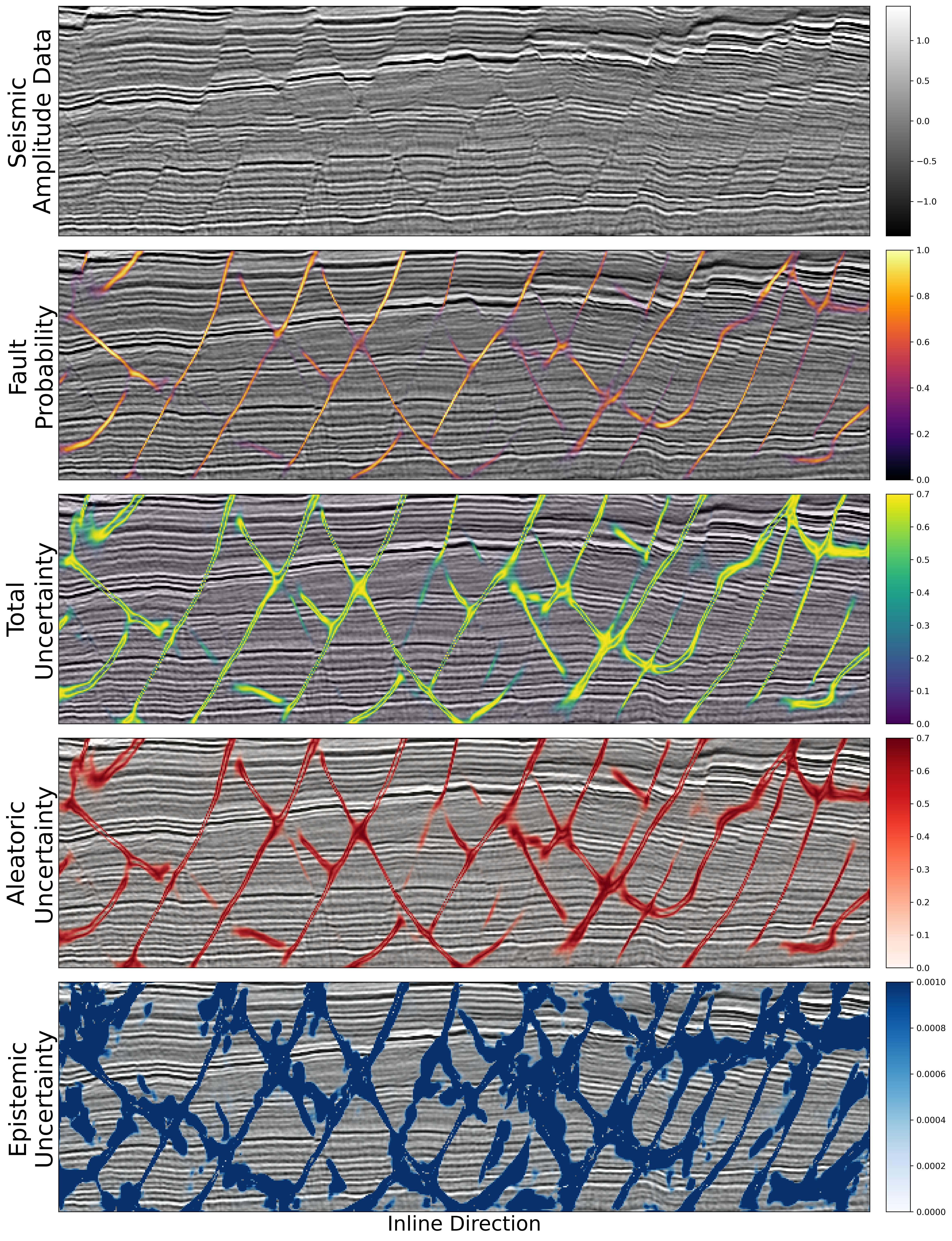} }}
\enspace
\subfloat[\centering Concrete Dropout]{{\includegraphics[width=5cm]{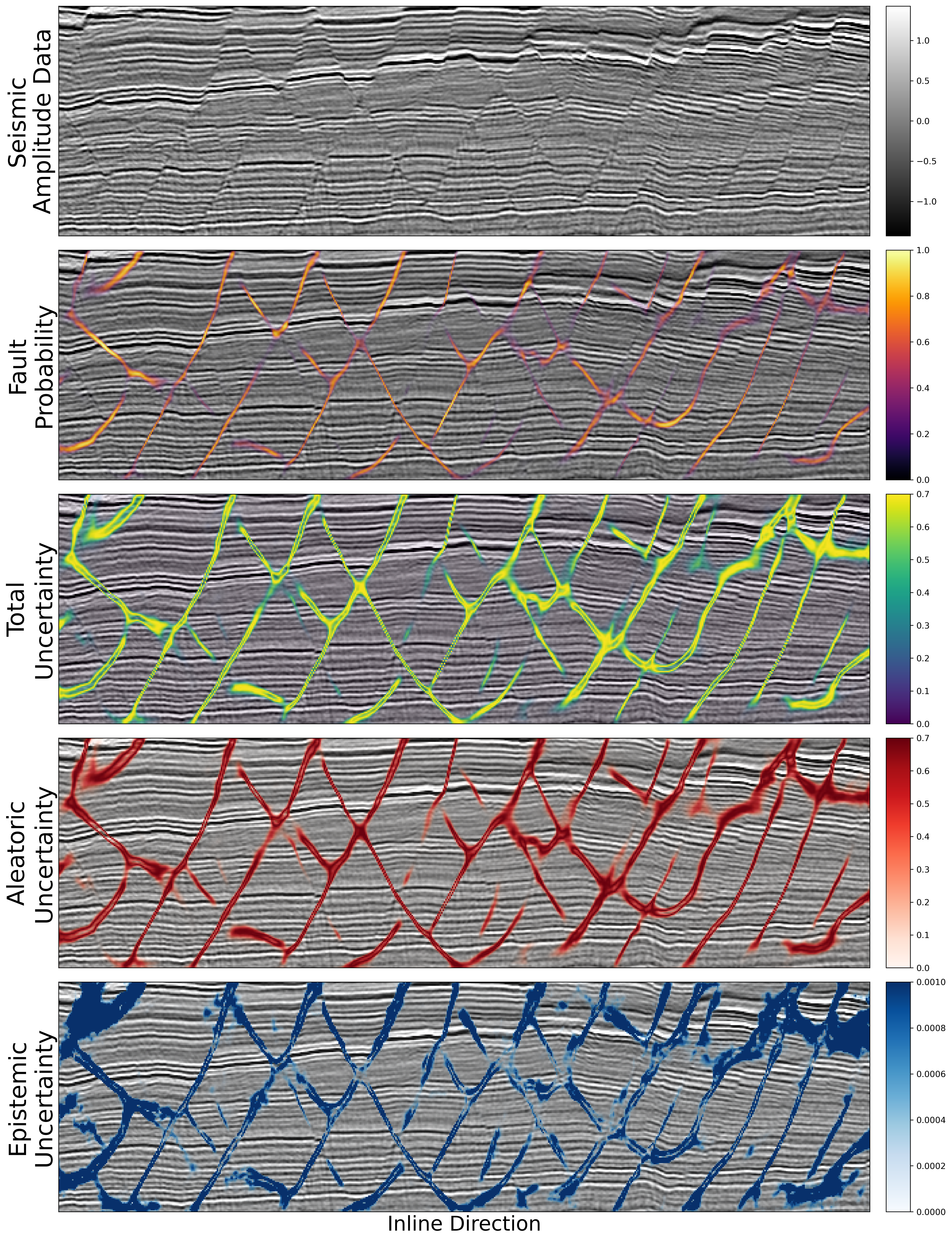} }}
\enspace
\subfloat[\centering SWAG]{{\includegraphics[width=5cm]{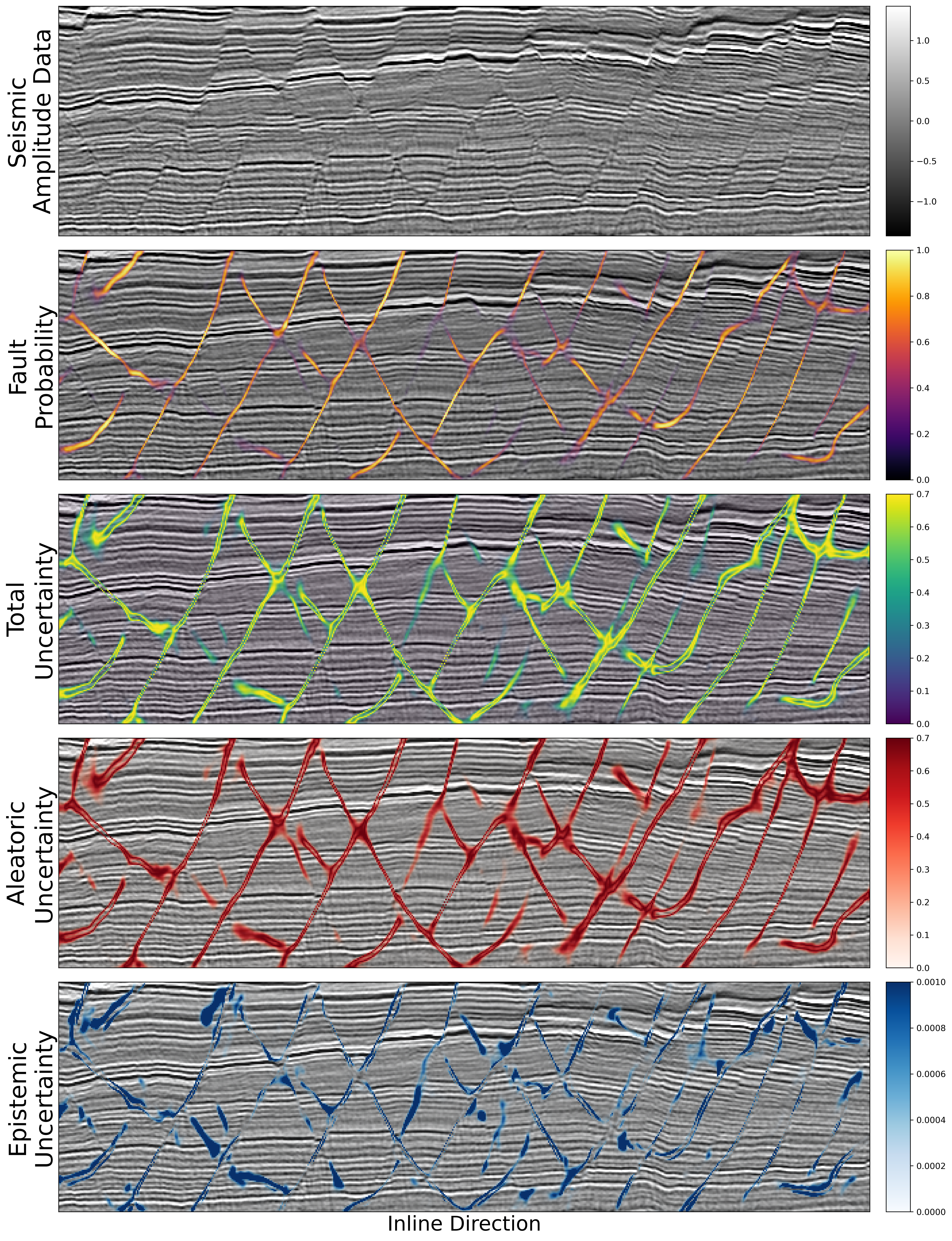} }}
\caption{Predicted probabilities and associated uncertainties on the SUN12 dataset from the North-Sea.}
\label{fig:sun12}
\end{figure*}
Based on synthetic data examples in the previous section, one can observe that the three proposed methods have the potential to provide calibrated probabilistic models together with associated uncertainties. Although there could be some challenges when dealing with out-of-distribution noise and artifacts, it was shown that these methods can still provide robust fault predictions with increased epistemic uncertainties compared to the noise-free base-case. Since the models were obtained using synthetic data only, making predictions on real seismic data has the assumption implicitly that the synthetic training data is representative of the distribution of the various real seismic datasets that the models will be evaluated on. Due to the possibility of out-of-distribution imaging artifacts and the diverse nature of natural faulting regimes, applying these models to real seismic data represents a challenging task that requires models to generalize from synthetic to real seismic data under possibly significant data distribution shift. 

In Figure~\ref{fig:sun12} we show results of applying the Deep Ensemble, Concrete Dropout, and SWAG models on the SUN12 dataset from the North Sea. We observe a well-defined fault system with minimal noise throughout the entire section. All three methods provide a well-defined fault probability attribute, with minimal differences observed between the three methods. When considering the total uncertainty, we can observe ring-like structures in the uncertainty attribute, with a low uncertainty close to high fault probabilities. This indicates that the network is highly certain about its own high probability predictions, but decreases away from the fault plane and at fault tips. Finally, we observe that the Deep Ensemble has the highest epistemic uncertainties, with the Concrete Dropout method and SWAG showing considerably lower values of the epistemic uncertainty. This reflects some of the observations seen on the evaluation performed on the synthetic datasets (Figure~\ref{fig:synthetic_mid_uncertainties}). Furthermore, this may indicate that the multiple modes captured by the Deep Ensemble may be critical for epistemic uncertainty estimation and that the local neighborhood of a mode explored by SWAG is smaller than for the case of Concrete Dropout models. Further refinement of the learning rate schedule for SWAG could increase the exploration of the local mode and hence lead to increased epistemic uncertainty estimates. 

\begin{figure*}
\centering
\subfloat[\centering Deep Ensemble]{{\includegraphics[width=5cm]{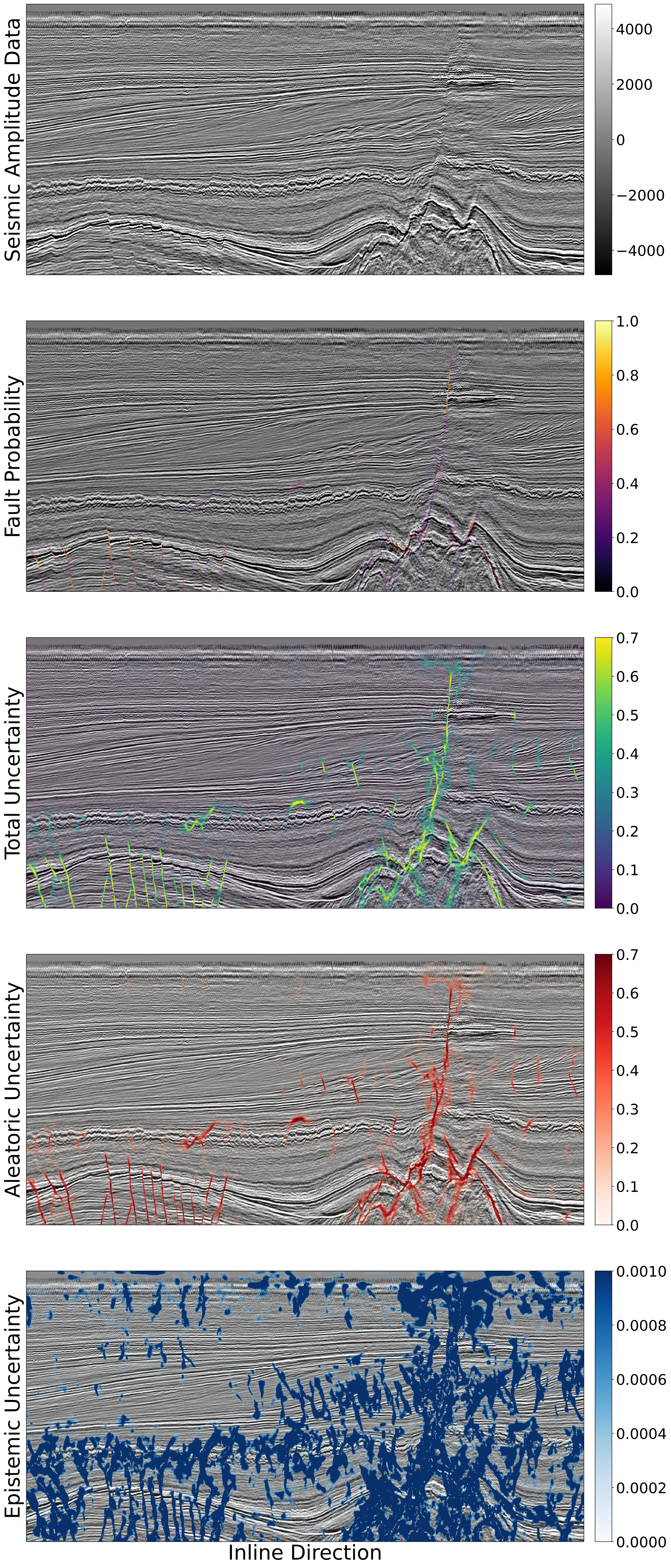} }}
\enspace
\subfloat[\centering Concrete Dropout]{{\includegraphics[width=5cm]{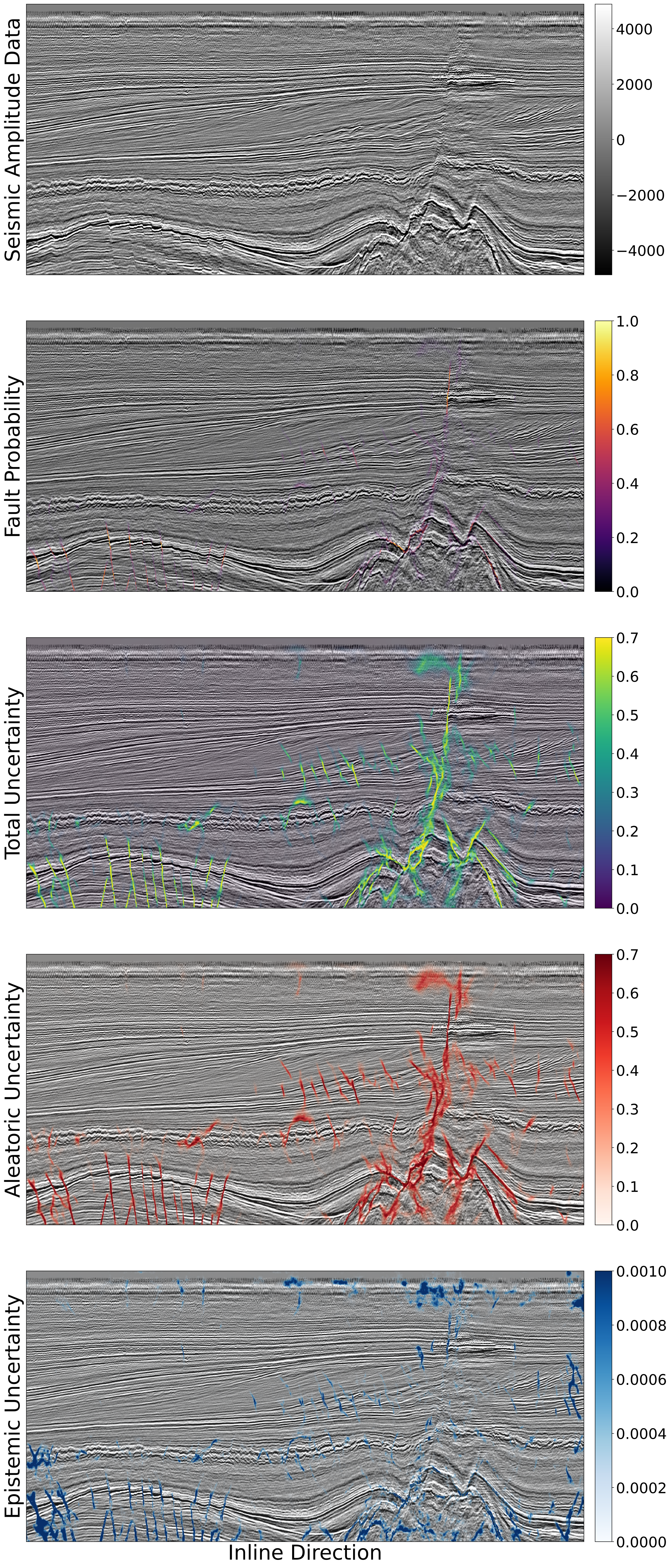} }}
\enspace
\subfloat[\centering SWAG]{{\includegraphics[width=5cm]{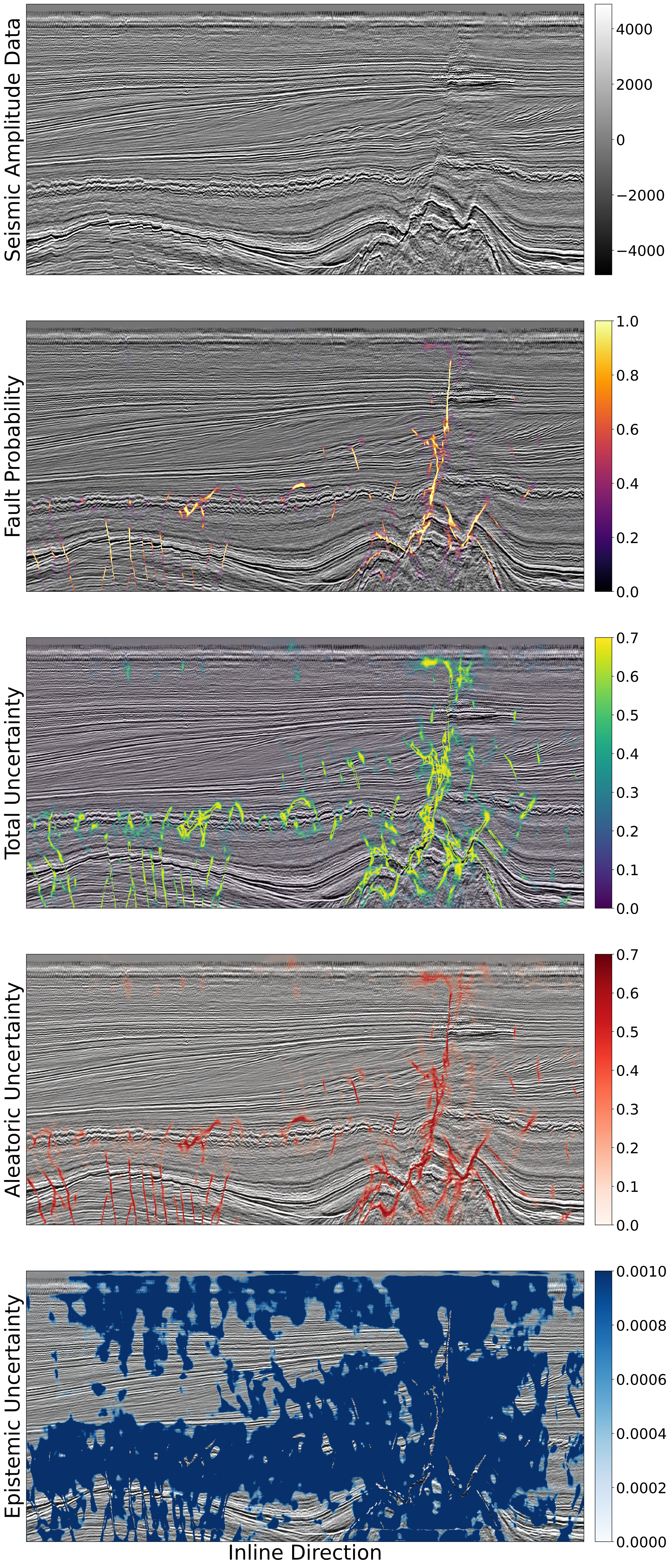}}}
\caption{Predicted probabilities and associated uncertainties on the dutch F3 dataset from the northern North-Sea.}
\label{fig:f3}
\end{figure*}

The dutch F3 dataset represents a seismic dataset that not only has a higher degree of coherent noise, but also a high diversity of fault geometries and discontinuous geological features such as the zone above the salt-dome located in the middle of the seismic inline of the F3 dataset shown in Figure~\ref{fig:f3}. The noisy shallow section and the salt dome represent areas where both the Deep Ensemble and SWAG methods show a very high degree of epistemic and aleatoric uncertainties, while the Concrete Dropout model only shows aleatoric uncertainty in both these areas of the seismic survey. The SWAG model shows a very well delineated fault probability attribute compared to the Deep Ensemble and the Concrete Dropout model. In this case, the Deep Ensemble and Concrete Dropout may be indicating a more conservative estimate of the true fault probabilities, which when the model is highly uncertain about its output should be close to values of 0.5, and hence the SWAG model could be considered overconfident in its own predictions. This would follow observations from the synthetic example (Table~\ref{tab:saltpepper5}) where the Deep Ensemble showed excellent calibration under distribution shift. All three methods indicate high aleatoric uncertainties close to the salt-dome, which again challenges the expression of this type of uncertainty by these approximate Bayesian methods and thus how one can interpret it in practice. Since this salt-dome and the associated seismic expression could be considered out-of-distribution for the synthetic training dataset, we would expect that the model shows a high epistemic uncertainty only, which here is not the case.

\begin{figure}
\centering
\subfloat[\centering Deep Ensemble]{{\includegraphics[width=7.5cm]{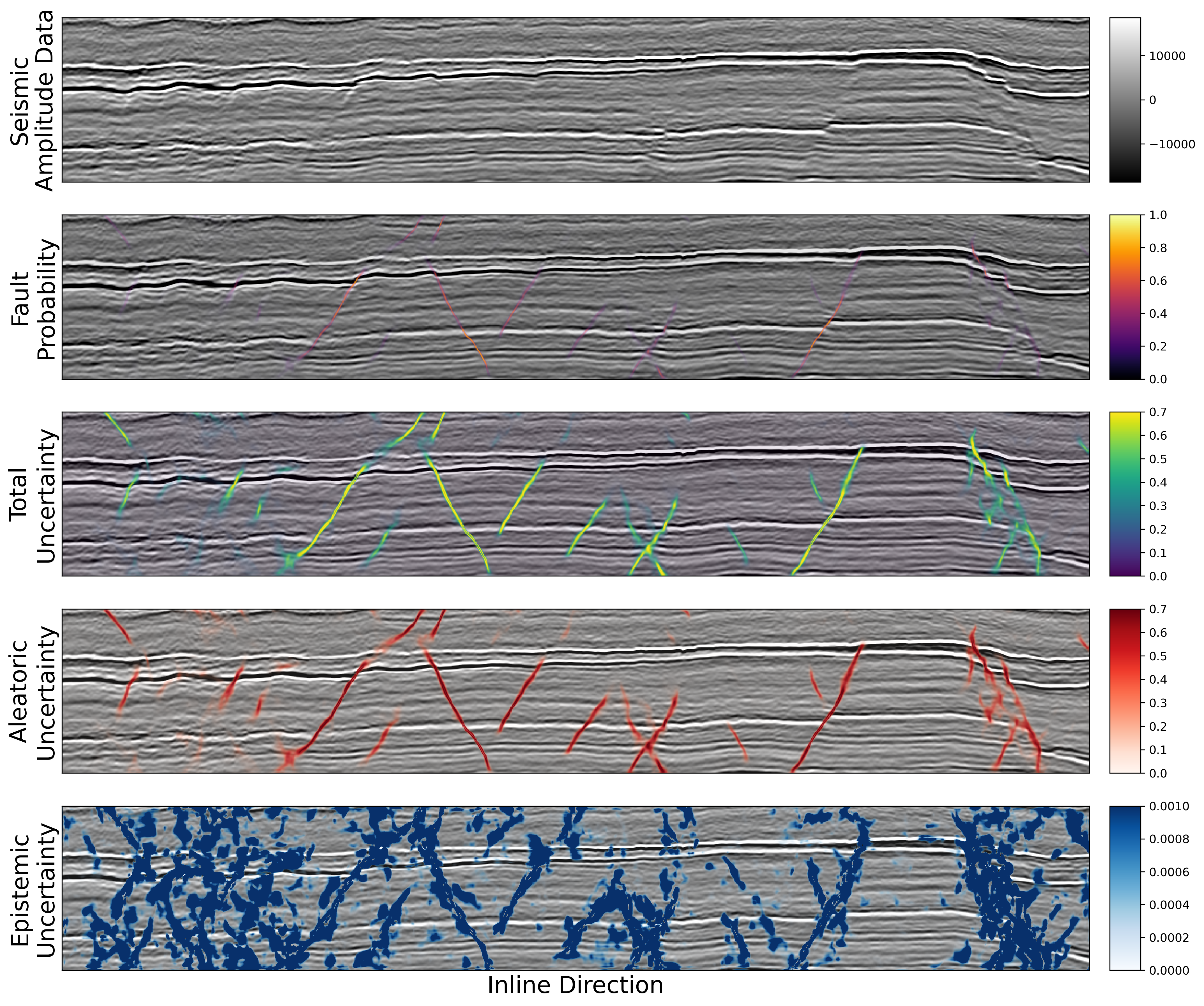}}}\\[-0.5ex]
\subfloat[\centering Concrete Dropout]{{\includegraphics[width=7.5cm]{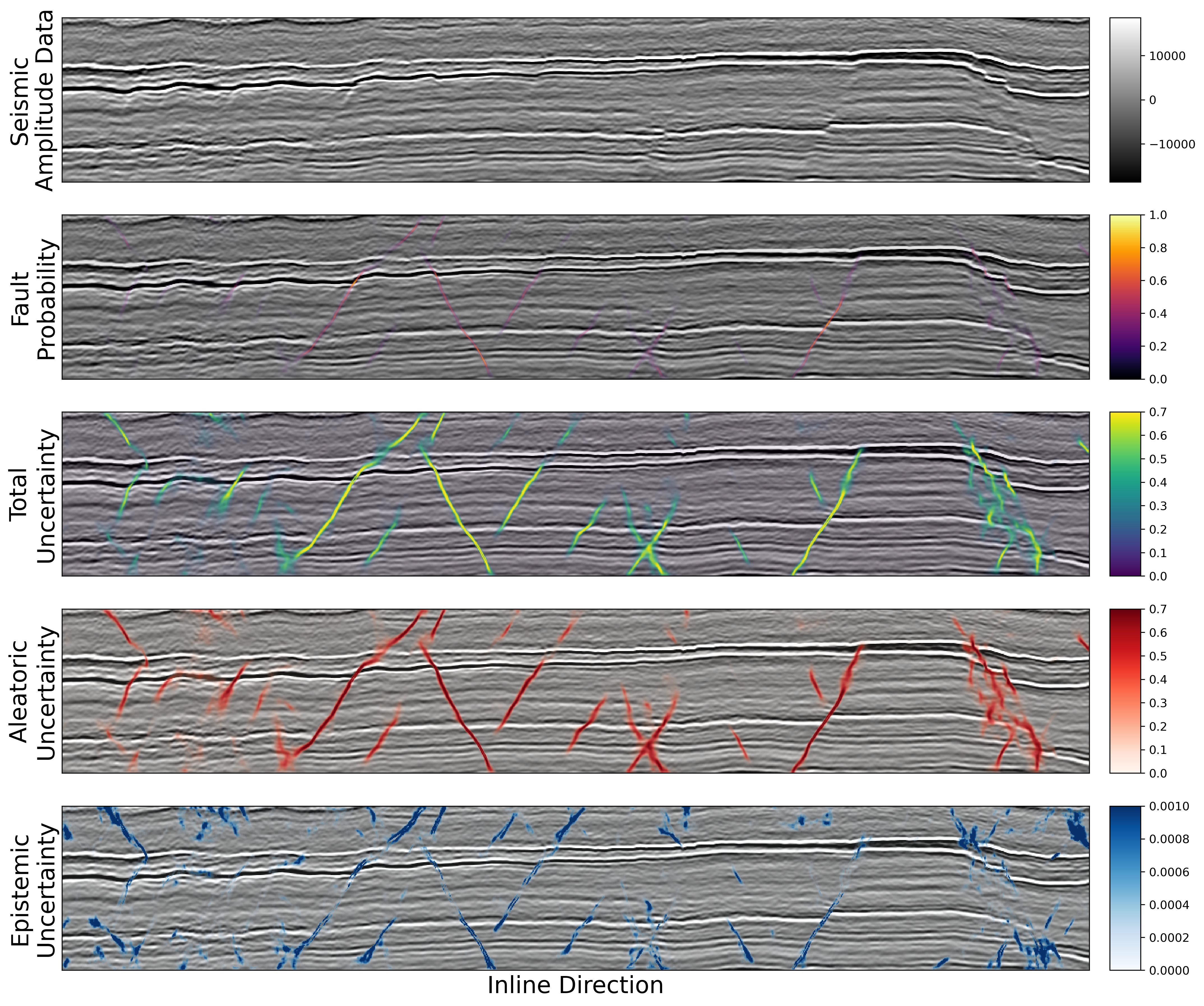}}}\\[-0.5ex]
\subfloat[\centering SWAG]{{\includegraphics[width=7.5cm]{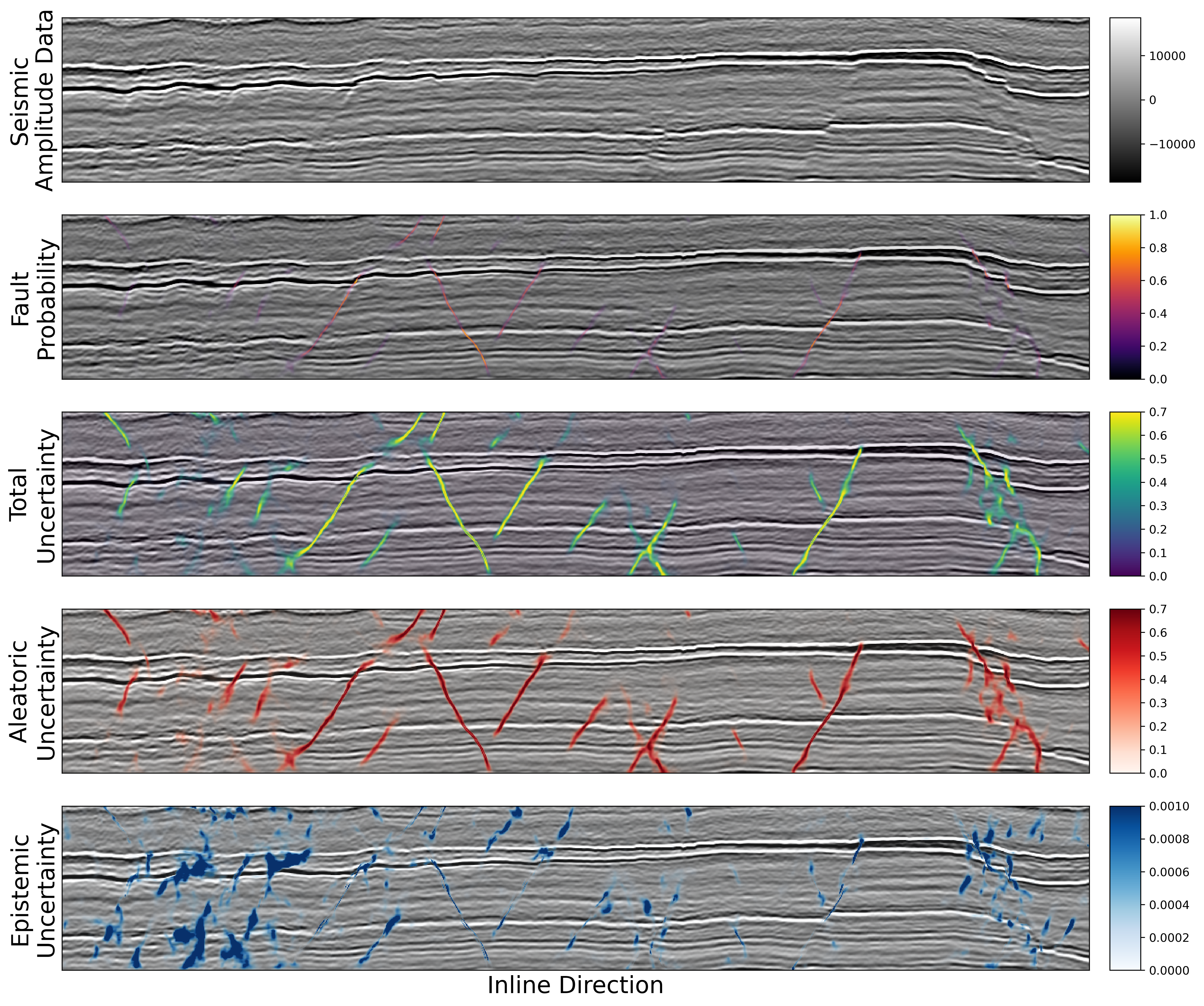}}}
\caption{Predicted probabilities and associated uncertainties on the Canning dataset from the North West Shelf Australia.}
\label{fig:canning}
\end{figure}
\vfill\null
Finally, we provide an example of an application of the three probabilistic methods on the Canning dataset from North West Shelf of Australia (Figure~\ref{fig:canning}). The faults in this dataset show a very subtle expression with small displacements where the fault plane is not imaged well. The fault probability clearly delineates regions where fault planes can be detected, with associated high total uncertainties around the fault planes and tips. Qualitatively we observe similar behavior to the SUN12 dataset (Figure~\ref{fig:sun12}) where the Deep Ensemble shows a larger degree of epistemic uncertainty compared to the Concrete Dropout and SWAG approaches. From this we can draw a number of conclusions: first, the seismic image qualities and artifacts of the Canning dataset may be closer to the data distribution of the SUN12 dataset, and therefore the data distribution of these two seismic datasets may be closer to the training datasets. Under this assumption, we can observe that the Concrete Dropout and SWAG method underestimate the amount of epistemic uncertainty associated with the corresponding models' predictions. Finally, the predictions and artifacts observed on the F3 dataset may occur because of a large distribution shift from the training dataset to the noisy seismic image quality of the F3 data. This of course warrants some caution to the practitioner who may rely on these uncertainty estimates, to ensure that the training data used also captures faulting and imaging modalities observed in the seismic data. Nevertheless, the high degree of epistemic uncertainty observed on the F3 dataset by the SWAG and the Deep Ensemble methods can be used to further refine the training dataset, by including additional imaging artifacts observed in the data into the training set and to provide guidance on which training data to provide to further improve these predictive models.

Additionally, we have to consider that the predictions of the fault probabilities on real seismic datasets are dependent on a number of factors such as the model architecture (Figure~\ref{fig:convnet}) and distribution of the model weights which can be considered as a prior, or as we have extensively shown, the choice of the approximate Bayesian inference method. 

One aspect to consider is the representation of the faults, the associated fault labels, and the seismic forward modeling process applied to the synthetic models. After convolution of the faulted geological model with a wavelet, some regions at the fault tips may not be uniquely detectable at the seismic resolution, whereas the labels are present up until the point where there is numerically zero displacements at the fault tips. The dataset and its labels may therefore imply that a fault is present where no fault can be detected based on the seismic image alone. Nevertheless, the fact that we have trained a model with these image and label configurations may lead the model to naturally overextend the faults because it is implicitly defined in the training set that faults may extend beyond the limits of seismic resolution. Therefore, it is important to consider the synthetic modeling process when applying these models since their results and the associated uncertainties will be a reflection of these modeling assumptions. When used with care, this can, in turn, be used to evaluate different geological and seismic modeling assumptions as these probabilistic models can be trained on different datasets and the impact of these assumptions can, in turn, be compared on the same seismic data. This is in a way the same principle that \citet{naeini2018machine} called "Machine Learning and Learning from Machines". 

\begin{figure*}
\centering
\includegraphics[width=\textwidth]{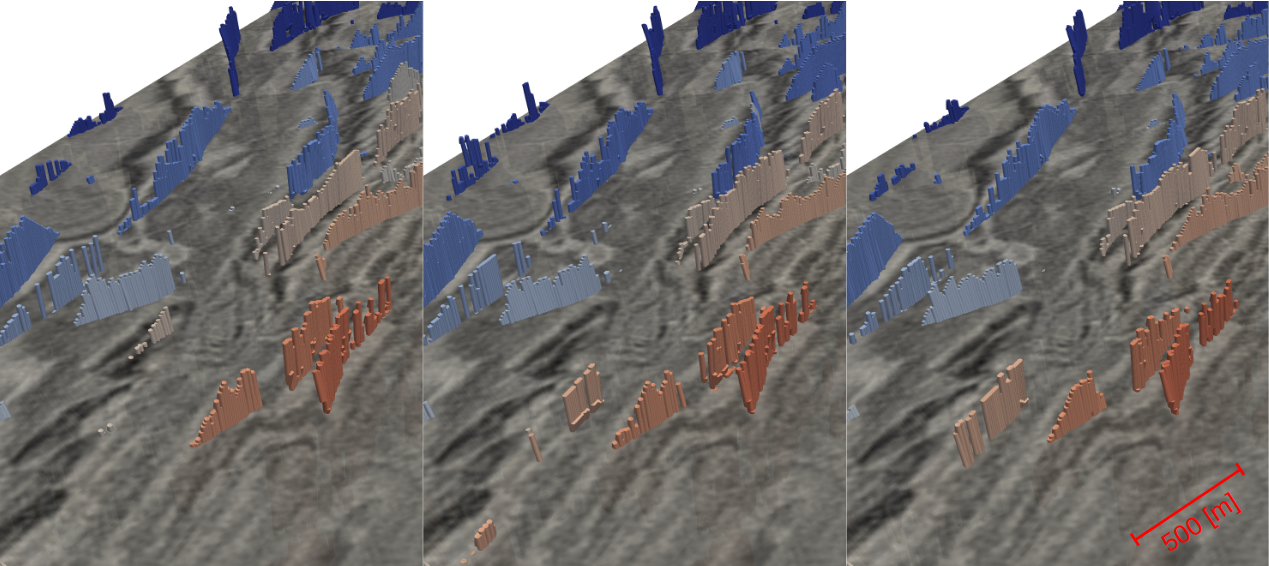}
\caption{Three realizations of the fault probability attribute and extracted fault sticks.}
\label{fig:fault_sticks}   
\end{figure*}

In practice, evaluation of a fault probability attribute is an intermediate step towards creating discrete surface representations of the faults located within a seismic interval of interest. These surfaces form the bounding planes of hydrocarbon or potential carbon capture and sequestration reservoirs. By considering each realization of the fault probability attribute sampled from the probabilistic models outlined in the \nameref{sec:methodology} section, we are able to also create realizations of discrete fault representations to provide further input into a probabilistic reservoir modeling workflow. In Figure~\ref{fig:fault_sticks} we show three realizations of the Concrete Dropout model where the fault probability attribute has been discretized into connected fault poly-lines which form the basis of fault-surface construction for a geological modeling workflow. The resulting fault network shows some degree of variance from one realization to the next, which in turn can have a significant impact on reservoir connectivity, history matching, and volumetric evaluation.

\section{Conclusions}\label{sec:conclusions}
We have presented three probabilistic approaches to obtaining probabilistic models of fault detection in seismic datasets using CNNs in a Bayesian formalism. The Deep Ensemble is a computationally expensive method that requires a full training run to be performed for each additional member of the ensemble. Conceptually the individual ensemble members represent modes of the posterior weight distribution (Equation~\ref{eq:posteriorpredictive}) and due to the multi-modal nature of this posterior distribution, numerous models have to be trained to capture the associated uncertainties in the predictive distribution. This is in many cases prohibitive to practical applications. 

Therefore we have considered two additional approaches; Concrete Dropout \citep{gal2017concrete} is a practical extension of the original Dropout method to approximate Bayesian CNNs \citep{gal2016dropout}. By optimizing the Dropout probabilities during the training process, there is no need for additional hyper-parameter tuning required to select appropriate probabilities for each Dropout operation in the neural network. This represents a significant reduction in the computational requirements compared to the original Dropout approach although there are additional cost to fine tune the length-scale parameter associated with the Concrete Dropout method. Because only a single model is trained, conceptually the Concrete Dropout approach explores the vicinity of a single mode of the posterior weight distribution. Finally, we presented an application of SWAG \citep{maddox2019simple} which is based around the equivalence of SGD training to Bayesian inference in probabilistic models \citep{mandt2017stochastic}. Stochastic samples of the posterior are used to form a low-rank Gaussian approximation of the posterior which characterizes the local neighborhood of a mode of the distribution.

Based on our results, we have shown that while for many quantitative metrics the Deep Ensemble outperforms Concrete Dropout and SWAG, both approaches provide competitive performance (Table~\ref{tab:metrics}), especially considering the reduced computational cost. All three methods perform well when considering out-of-distribution noise artifacts that are commonly found in seismic datasets (Table~\ref{tab:saltpepper5}). It is also possible to explore other modes of the posterior using a hybrid combination of Deep Ensemble and SWAG or Concrete Dropout to improve the quality of uncertainty estimation and calibration \citep{wilson2020bayesian}.

Furthermore, we show that using an unbalanced BCE-loss is sufficient for training deterministic and probabilistic models on a synthetic dataset of seismic images providing well-calibrated predictions. Using a balanced loss \citep{wu2019faultseg3d} leads to faster convergence but overall miscalibrated probability distributions with a high number of false-positive predictions. In this case, additional post-processing methods as presented by \citet{guo2017calibration} could be applied to improve the calibration of these models.

Our applications of these Bayesian CNNs on real seismic datasets from numerous Basins around the world (Figure~\ref{fig:sun12}-\ref{fig:canning}) show that high-quality fault probability and uncertainty attributes can be obtained. Qualitatively we find that the aleatoric and epistemic uncertainties are not easily disentangled as was extensively discussed with synthetic and real data examples. This could be due to a shift in the data distribution from the training to the real seismic datasets which can possibly lead to misappropriation of the respective uncertainties. 

An example of such behavior can be seen on the F3 dataset (Figure~\ref{fig:f3}) where the salt dome can be considered a feature not contained in the training dataset, and hence this should induce high epistemic uncertainties, yet we also observe high aleatoric uncertainties which can be attributed to irreducible model uncertainties. This warrants further investigation into how to characterize the two types of uncertainties in geophysical applications. Nevertheless, the epistemic uncertainties obtained from these models provide guidance on where to select additional training data to provide further improvements in the model's abilities to make predictions of these features. 

Finally, we have shown that using the obtained stochastic realizations of the fault attributes as samples from the posterior predictive distribution can be used to assess the impact of such uncertain nature of this type of interpretation on subsequent reservoir characterization workflows. The extracted discrete fault representations obtained different realizations of the fault probability attributes. The resulting fault geometries could in turn produce diverse connectivity patterns in a reservoir interval and hence can represent different geological scenarios that need to be considered in an uncertainty-based characterization and decision-making workflow. The presented methods are agnostic to the specific geoscience application of deep neural networks and can be readily incorporated into other applications such as property prediction approaches \citep{mosser19uncertainty}. 

\section{Acknowledgments}
We would like to thank Earth Science Analytics for permission to publish this work. We thank the Netherlands Organization for Applied Scientific Research for releasing the F3 seismic dataset. We further would like to thank Geoscience Australia and ConocoPhillips Ltd. for providing access to the Canning dataset.
We acknowledge the Diskos NDR (diskos.no) managed by the Norwegian Petroleum Directorate for access to the public seismic data volumes. We would like to thank Andrew Gordon Wilson, Pavel Izmailov, and Wesley Maddox for providing open-source reference implementations of SWAG

\bibliography{bib}
\end{document}